\documentclass[12pt]{elsarticle}
\usepackage{geometry}
\usepackage{float}
\usepackage{booktabs}
\usepackage{caption}
\usepackage{xcolor}
\usepackage{multirow}
\usepackage{tikz}
\usetikzlibrary{positioning, shapes, decorations.markings, decorations.pathreplacing, decorations.pathmorphing, arrows, calc, matrix}
\usepackage{comment}
\usepackage{enumerate}
\usepackage{tabto}
\NumTabs{6}
\usepackage{calligra}
\usepackage{url}
\usepackage{subfig}
\usepackage{chngcntr}
\usepackage{amsmath,amssymb,amsfonts,amsthm,mathtools}
\usepackage{lineno}

\usepackage{cleveref}
\def\appendixname{}
\newcommand{\abs}[1]{\left\vert#1\right\vert}
\newcommand{\E}{\mathop{{}\mathbb{E}}}
\newcommand*\mean[1]{\overline{#1}}
\newcommand{\vect}[1]{\boldsymbol{#1}}
\DeclareMathOperator*{\argmin}{arg\,min}

\journal{}

\begin{document}

\input{diagram_influence.bst}
\input{diagram_nn.bst}
\input{diagram_mcst.bst}

\begin{frontmatter}



\title{\large An investigation of belief-free DRL and MCTS for inspection and maintenance planning}


\author[inst1]{Daniel Koutas}
\ead{daniel.koutas@tum.de}

\author[inst1]{Elizabeth Bismut}
\ead{elizabeth.bismut@tum.de}
\author[inst1]{Daniel Straub}
\ead{straub@tum.de}

\affiliation[inst1]{organization={Engineering Risk Analysis Group}, addressline={Technische Universität München}, country={Germany}}

\begin{abstract}
We propose a novel Deep Reinforcement Learning (DRL) architecture for sequential decision processes under uncertainty, as encountered in inspection and maintenance (I\&M) planning. Unlike other DRL algorithms for (I\&M) planning, the proposed +RQN architecture dispenses with computing the belief state and directly handles erroneous observations instead. We apply the algorithm to a basic I\&M planning problem for a one-component system subject to deterioration. In addition, we investigate the performance of Monte Carlo tree search for the I\&M problem and compare it to the +RQN. The comparison includes a statistical analysis of the two methods' resulting policies, as well as their visualization in the belief space.
\end{abstract}



\begin{keyword}
one-component deteriorating system \sep maintenance planning \sep partially observable MDP \sep deep reinforcement learning \sep neural networks \sep Monte Carlo tree search; 
\end{keyword}

\end{frontmatter}

%\linenumbers


%%       \citet{<label>} ==> Jones et al. [21]
%%       \citep{<label>} ==> [21]
%%
%
%\newacronym{drl}{DRL}{Deep Reinforcement Learning}
%\newacronym{i&m}{I\&M}{Inspection and Maintenance}
%\newacronym{+rqn}{+RQN}{Action-specific Deep Dueling Recurrent Q-network}
%\printglossary[type=\acronymtype]

\section*{Abbreviations}
\begin{comment}
\begin{tabular}{l l}
    \textbf{+RQN} & Action-specific Deep Dueling Recurrent Q-network \\
    \textbf{DCMAC} & Deep Centralized Multi-agent Actor Critic \\
    \textbf{DDMAC} & Deep Decentralized Multi-agent Actor Critic \\
    \textbf{DRL}   & Deep Reinforcement Learning \\
    \textbf{DQN}  & Deep Q-Network \\
    \textbf{DDQN} & Double Deep Q-Network \\
    \textbf{FC} & Fully Connected \\
    \textbf{I\&M}  & Inspection and Maintenance \\
    \textbf{LCC} & Life Cycle Cost \\
    \textbf{LSTM} & Long Short-Term Memory \\
    \textbf{MC} & Monte Carlo \\
    \textbf{MCTS} & Monte Carlo Tree Search \\
    \textbf{MDP}  & Markov Decision Process \\
    \textbf{MSE} & Mean-Squared Error \\
    \textbf{NN} & Neural Network \\
    \textbf{POMDP} & Partially Observable Markov Decision Process \\
    \textbf{RL} & Reinforcement Learning \\
    \textbf{RV} & Random Variable \\
    \textbf{UCT} & Upper Confidence Bound for Trees \\
    \textbf{VI} & Value Iteration
\end{tabular}
\end{comment}

\noindent
\textbf{+RQN}   \tab Action-specific Deep Dueling Recurrent Q-network 

\noindent
\textbf{DCMAC}  \tab Deep Centralized Multi-agent Actor Critic

\noindent
\textbf{DDMAC}  \tab Deep Decentralized Multi-agent Actor Critic

\noindent
\textbf{DRL}    \tab Deep Reinforcement Learning

\noindent
\textbf{DQN}    \tab Deep Q-Network

\noindent
\textbf{DDQN}   \tab Double Deep Q-Network

\noindent
\textbf{FC}     \tab Fully Connected

\noindent
\textbf{I\&M}   \tab Inspection and Maintenance

\noindent
\textbf{LCC}    \tab Life Cycle Cost

\noindent
\textbf{LSTM}   \tab Long Short-Term Memory

\noindent
\textbf{MC}     \tab Monte Carlo

\noindent
\textbf{MCTS}   \tab Monte Carlo Tree Search

\noindent
\textbf{MDP}    \tab Markov Decision Process

\noindent
\textbf{MSE}    \tab Mean-Squared Error

\noindent
\textbf{NN}     \tab Neural Network

\noindent
\textbf{POMDP}  \tab Partially Observable Markov Decision Process

\noindent
\textbf{RL}     \tab Reinforcement Learning

\noindent
\textbf{RV}     \tab Random Variable

\noindent
\textbf{UCT}    \tab Upper Confidence Bound for Trees

\noindent
\textbf{VI}     \tab Value Iteration

\section{Introduction}
\label{sec:Introduction}
Reliable civil infrastructure, such as power, water and gas distribution systems or transportation networks, is essential for society. Large efforts are therefore spent on properly maintaining these systems. However, at present such maintenance is based mainly on simple legacy rules, such as fixed inspection intervals, combined with expert judgement.
% provides a wide range of essential services for the productivity of a country and the quality of life of individuals. Neglecting proper maintenance imposes costs on users in the short run (e.g. costs of repairing a car due to bad roads) and greater costs due to failure and rebuilding in the long run.
There is a significant potential for optimal inspection and maintenance (I\&M) planning that makes best use of the information at hand to ensure safe and reliable infrastructure while being sustainable and cost-efficient \citep{rioja2013valueofinfrastructure,daniela2011decision,frangopol2004probabilistic}.

I\&M planning is a sequential decision making problem under uncertainty. One challenge in deriving optimal I\&M decisions is the presence of large epistemic and aleatoric uncertainties associated with the system properties, load, representation model, and measurements \citep{bismut2021optimal,StraubLectures,sullivan2015introduction,madanat1993optimal}. %\citep{StraubLectures,sullivan2015introduction,smith2013uncertainty,madanat1993optimal}. 
Another major challenge is the exponential increase in possible I\&M strategies with the number of components and the considered time horizon \citep{bismut2021optimal,luque2019risk}.
Standard practice for dealing with these challenges is the use of established decision heuristics, e.g., safety factors during design, predetermined scheduled inspections, and threshold- or failure-based replacement of components \citep{melchers2018structural,rausand2003system,isreportcard}. The parameters of these heuristics can then be optimized to find good I\&M strategies \citep{luque2019risk,bismut2021optimal}. However, heuristics can be suboptimal and finding good heuristics is challenging. 

Another approach to embed uncertainty into the inherently sequential nature of inspection and maintenance problems, is to integrate probabilistic models into decision process models \citep{yuen2010bayesian,kim2013generalized,kim2011probabilistic}. Under certain conditions, these sequential decision problems under uncertainty can be modeled as Partially Observable Markov Decision Processes (POMDPs), which provide an efficient framework for optimal decision making, and can additionally account for measurement errors \citep{kochenderfer2015decision,andriotis2021deep,kaelbling1998planning}. The POMDP is in general intractable \citep{papadimitriou1987complexity}. Many approaches for solving the POMDP use the belief state representation, which incorporates the entire information, i.e., actions and observations up to the current point \citep{meng2021memory,andriotis2019managing,schobi2016maintenance,corotis2005modeling,kochenderfer2015decision}. However, these methods require an explicit probabilistic model of the environment to calculate the transition probabilities between states as well as the belief states, which is not always available. In addition, they typically are not computationally efficient beyond small state and action spaces \citep{meng2021memory}. This hinders their application to I\&M planning of infrastructure systems, where the investigated systems are usually consisting of a larger number of components.

Reinforcement learning approaches to solve POMDPs have gained in popularity, including Deep Reinforcement Learning (DRL) with neural networks (NNs), and Monte Carlo Tree Search (MCTS). There exist numerous variants of NNs for discrete \citep{hausknecht2015drqn,lample2017playingfps,zhu2017action-specific} and continuous \citep{song2018recurrent,wang2019autonomous,duan2016benchmarking} action space control, employing for example Deep Q-networks (DQNs) \citep{mnih2013playingatari,mnih2015human}, Double DQNs (DDQNs) \citep{brim2020ddqn,lv2019stochasticddqn} or actor-critic architectures \citep{haarnoja2018softactorcritic}. 
Although MCTS was originally formulated for fully observable domains with great success \citep{silver2016mastering,silver2018general}, it has also been applied to POMDPs \citep{silver2010pomcp,katt2017learning}. 

Both NNs and MCTS have been heavily researched in the field of computer games, which provide a safe (i.e., no real-life consequences) and controllable environment with a variety of complex problems to solve (2D, 3D, single-agent, multi-agent, etc.) with an infinite supply of useful data that is much faster than real-time \citep{shao2019gamesurvey}. The success of these methods in this application has motivated researchers to apply them to I\&M planning \citep[e.g.,][]{andriotis2019managing,andriotis2021deep,zhou2022reinforcement,saifullah2022deep}. However, this problem's specific characteristics e.g., sparse rewards due to low probability of failure, can pose a challenge to DRL methods, the efficiency of which remains to be systematically assessed.

The literature on solving POMDPs with DRL in the context of I\&M is fairly limited. Most studies have focused on fully observable MDPs, for instance coupling Bayesian particle filters and a DQN for real-time maintenance policies \citep{skordilis2020deep}, employing a DDQN for preventive maintenance of a serial production line \citep{huang2020deep}, coupling a pre-trained NN for reward estimation with a DDQN for maintenance of multi-component systems \citep{nguyen2022artificial}, and adopting a DDQN for rail renewal and maintenance planning \citep{mohammadi2022deep}. Concerning POMDPs, \citet{andriotis2019managing} developed the Deep Centralized Multi-agent Actor Critic (DCMAC) architecture for multi-component systems operating in high-dimensional spaces, with extended applications for roadway network maintenance \citep{zhou2022reinforcement}. The corresponding decentralized version (DDMAC), where each agent has a separate policy network \citep{andriotis2021deep}, has been applied to life cycle bridge assessment \citep{saifullah2022deep} and 9-out-of-10 systems \citep{morato2022inference}. However, both DCMAC and DDMAC take the belief state of the system as an input, which is in general computationally expensive to obtain for a system with many components and arbitrary state evolution processes. Thus, newer studies \citep[e.g.,][]{giacomo,hettegger} have shifted the focus to observation-based DRL. However, a problem setting concerning continuous state and continuous erroneous observations has not been considered, yet.   

In a similarly limited manner, MCTS has been applied to maintenance planning problems modeled as MDPs. Examples with MCTS include, for instance, finding stochastic schedules in active distribution networks \citep{shang2020stochastic}, in combination with genetic algorithms for condition-based maintenance \citep{hoffman2021online}, or combined with NNs for wind turbine maintenance \citep{holmgren2019general}. To the best of our knowledge, MCTS has not been applied to POMDPs in the context of I\&M.

The purpose of this paper is twofold. Firstly, we propose a DRL architecture for POMDP and I\&M planning, which does not require the computation of the belief state. The proposed NN combines the features of the Action-specific Deep Recurrent Q-Network \citep{zhu2017action-specific} and the dueling architecture \citep{wang2016dueling}. The resulting +RQN architecture is able to deal directly with erroneous observations over the whole life cycle of the system.

Secondly,  we investigate the performance of MCTS when applied to I\&M planning. In this context, we perform a systematic comparison of the proposed +RQN and MCTS. The investigated problem is a one-component system subject to deterioration and is formulated as a POMDP, for which an exact solution is available, because of linear Gaussian assumptions for the model dynamics. The analysis includes a comparison of performance, i.e., the achieved optimized expected life cycle costs (LCC) and the computation time. It is carried out for different measurement errors.  We also review the information carried by two metrics to compare the resulting policies of the two methods, namely via a statistical analysis and a visualization in the belief space. The solutions from both methods are compared to the exact POMDP solution.

The structure of the paper is as follows. \Cref{sec:Problem} introduces the investigated problem as well as sequential decision making along with the key definitions and metrics needed for the employed RL methods. \Cref{sec:NNs} explains the workings of the NN architecture used herein, and \Cref{sec:MCTS} illustrates how the MCTS method has been adapted for solving the proposed problem. \Cref{sec:Metric_for_comparison} is dedicated to the metrics we employ to compare the NN and MCTS solutions, and \Cref{subsec:computation_time,subsec:numerical_performance,subsec:policy_comparison} contain the respective results. \Cref{sec:Discussion} discusses the obtained solutions and policies, and gives insight into the advantages and disadvantages of the two approaches.
%
%
%
%
%%%%%%%%%%%%%%%%%%%%%%%%%%%%%%%%%%%%%%%%%%%%%%%%%%%%
%%%%%%%%%%%%%%%%%%%%%%%%%%%%%%%%%%%%%%%%%%%%%%%%%%%%
%%%%%%%%%%%%%%%%%%%%%%%%%%%%%%%%%%%%%%%%%%%%%%%%%%%%
%
%
%
%
%
\section{Basic maintenance problem}
\label{sec:Problem}
\subsection{Investigated system}
\label{subsec:investigated_system}

For the numerical investigations in this paper,
we study a one-component system subject to 
deterioration, taken from \citep{Noichl}. It is 
modeled with two random variables (RVs): $D$ 
representing the \emph{deterioration state} and 
$K$ representing the \emph{deterioration rate}. 
The subscript $t$ indicates timesteps, where
$t=0,~1,~2,~...,~T_{\mathrm{end}}$, with finite 
time horizon $T_{\mathrm{end}}$. The generic 
deterioration model is given as
\begin{linenomath}
\begin{equation}
\label{eq_additive_model}
    D_t = D_0 + t \cdot K_0   \qquad \Longleftrightarrow \qquad
    \left\{
    \begin{array}{l}
        D_t = D_{t-1} + K_{t-1}  \\[0.2em]
        K_t = K_{t-1}
    \end{array}
    \right.,
\end{equation}
\end{linenomath}
where $D_0$ and $K_0$ are normally distributed and independent. \Cref{eq_additive_model} shows that the deterioration process is modeled as a Markov process through state space augmentation. The deterioration $D_t$ is observable with a Gaussian measurement noise, through the measurement random variable $O_t$, i.e., $O_t\sim \mathcal{N}(D_t,\sigma_E)$. 

Four actions $a_0 - a_3$ are available for counteracting the deterioration and ultimately the failure of the structure. The action $A_t$ is taken after observation $O_t$ and affects $D_{t+1}$ and/or $K_{t+1}$ (see \Cref{sec:app_model_info}). The effects of the actions on the system are detailed in  \Cref{tab:action_effect_on_samples_and_belief}. The structure fails when the deterioration exceeds the critical deterioration $d_{cr}$.
In the failed state, an annual failure cost is incurred until the system is either repaired or replaced (no automatic setback of the system to the initial state). In addition, each action $a_i$ has a specific cost $c_{a_i}$ incurred at time $t$. 

\begin{comment}
The goal is to find a policy which minimizes the sum of total expected discounted cost in \Cref{eq_expected_LCC}, i.e., the sum of the expected action costs and the expected failure costs \citep{Noichl}.

The deterioration increases steadily
until the critical deterioration $d_{cr}$ is reached and the structure fails. 

The structure fails when the deterioration exceeds the critical deterioration $d_{cr}$.
It stays in the failed state until it is repaired,
replaced or the end of the system lifetime is reached.
\end{comment}

\Cref{fig:complete_influence_diagram} depicts the generic influence diagram of the corresponding POMDP.
%
%\snapshotid
%
\completeid

This case study is set up such that linearity, and hence also the normality of any set of RVs, is conserved (see \Cref{tab:action_effect_on_samples_and_belief}). As a result, the belief state and all transitions of the belief-MDP can be computed analytically. 

Moreover, in our case, the covariance matrix does not depend on the observations and the actions taken, and can hence be pre-computed for all timesteps. Thus, the actions and observations only influence the prior and posterior means of $D_t$ and $K_t$, respectively (see \Cref{sec:app_model_info}).

The model assumption allows for the system to regenerate if $K_t$ is negative. However, 1) we set up the numerical values so that we limit this effect, 2) it is a useful assumption for obtaining a reference solution and 3) the solution methods introduced hereafter do not require it.

%
\begin{comment}
%
The posterior means are given in \Cref{sec:app_model_info} \Cref{subsubsec:app_posterior_mean_values} :
%
\begin{align}
\label{eq_mu_D_t_posterior}
\mu_{D, t}'' &=\frac{\sigma_{D, t}^{\prime \prime 2}}{\sigma_{E}^{2}} O_{t}+\frac{\sigma_{D, t}^{\prime \prime 2}}{\sigma_{D, t}^{\prime 2}} \mu_{D, t}' 
\\[0.5em]
\label{eq_mu_K_t_posterior}
\mu_{K, t}'' &=\frac{\rho_{t}' \sigma_{D, t}' \sigma_{K, t}'}{\sigma_{E}^{2}+\sigma_{D, t}^{\prime 2}}\left(O_{t}-\mu_{D, t}'\right)+\mu_{K, t}',
\end{align}
whereas the effect of the actions on the prior means is given in \Cref{tab:action_effect_on_samples_and_belief}.
%
\end{comment}

%
\begin{comment}
%
\begin{itemize}
    \item one-component system subject to deterioration with failure threshold, discrete time, observation variable, 4 actions with associated costs available, finite horizon
    \item show basic evolution equation and influence diagram
    \item show table with actions and all costs, table with specific values for model parameters in 
    \item everything normally distributed at every timestep, beliefs analytically available, standard deviations and correlation independent of observation values (-> can be precomputed)
    \item belief formulas and action effects on belief in 
    \item include benchmark LCCs?
\end{itemize}
%
\end{comment}
%
%
%
%%%%%%%%%%%%%%%%%%%%%%%%%%%%%%%%%%%%%%%%%%%%%
%%%%%%%%%%%%%%%%%%%%%%%%%%%%%%%%%%%%%%%%%%%%%%
%
%
\subsection{Sequential Decision Making}
\label{subsec:POMDPs}
At every timestep, the operator has to decide which action to choose based on the history of observations and actions; hence they try to solve a sequential decision making problem. Specifically, as the deterioration state $D_t$ is only observable through erroneous measurements $O_t$, and the deterioration rate $K_t$ is not observable at all, the investigated setup falls under the category of a Partially Observable Markov Decision Processes (POMDP) \citep{kochenderfer2015decision}. One can transform a POMDP into a belief MDP by replacing the states with the belief (vector) as the variable of interest, and then employ conventional methods for solving MDPs, such as value iteration (VI) or policy iteration \citep{braziunas2003pomdp}. We utilize this belief state representation to obtain a reference solution for the numerical investigations (see \Cref{subsec:POMDP_reference_solution,sec:Results}). 
However, the focus of this paper is specifically on reinforcement learning (RL) techniques that can directly deal with observation-action sequences and hence do not need the belief state representation. 

The goal is to find a sequence of actions that minimizes the expected life cycle cost (LCC), which is defined as the sum of discounted expected action and failure costs:
\begin{linenomath}
\begin{equation}
    \label{eq_expected_LCC}
    \mean{\mathrm{LCC}} =    \E \left[\mathrm{LCC}\right] = 
    \sum_{t=0}^{T_{end}} \gamma^t \cdot 
    \E \left[ C(A_t) + C(F_t) \right].
\end{equation}
\end{linenomath}
In standard literature, the two costs associated with action and failure are summarized in a single cost $C(s,a)$, which is the immediate cost resulting from executing action $a$ in state $s$ of the system. Hence, will adopt this notation in the following. 

The decision-making rule, which determines the action to take in function of the available information, is called the \emph{policy} $\pi$. In general, the policy is time- and history-dependent \citep{kochenderfer2015decision, dong2020deep}. There exists a mapping from the current observation-action history $h_t=(o_{1:t}, a_{1:t-1})$ to the time-agnostic belief over the set of system states $b(s_t)=p(s_t|o_{1:t},a_{1:t-1})$, where $b(s)$ represents the probability of the system being in state $s$, when the agent's belief state is $b$ \citep{cassandra1994acting}. Hence, the policy as well as other functions can be expressed in terms of both:
\begin{linenomath}
\begin{equation}
\label{eq_policy_def}
    \pi = \pi_t(h_t) = \pi_t(b).
\end{equation}
\end{linenomath}
Accordingly, the ideal policy $\pi^{*}$ determines the ideal action to take to reach the set goal. For finite-horizon problems (as for our case study), $\pi^{*}$ is generally time-dependent. In our case, the set of ideal policies $\{\pi^{*}_t,~t=1,~2,~...,~T_{\mathrm{end}}-1\}$ is the one that minimizes $\mean{\mathrm{LCC}}$. To find an expression for $\pi^{*}_t$, we substitute the global LCC measure (\Cref{eq_expected_LCC}) with recursively defined value functions.

\begin{comment}
@@ this is unclear. it sounds like the same policy is used for every time after t, which however would not be optimal. @@
%
%
%
@@ There is a disconnect between this paragraph and the above Eq. 6. In both cases, an optimization over $\pi$ is performed, so how are these connected? 
(This includes the relationship between LCC and C.) @@
\end{comment}

A \emph{state value function} assigns a value to a particular (belief) state at a specific point in time. We denote with $V^{\pi}_t(b)$ the sum of expected discounted costs when following policy $\pi$ starting from belief $b$ at time $t$ \citep{walraven2019point}. The optimal value function is then defined as \citep{cassandra1994acting}:
\begin{linenomath}
\begin{align}
\label{eq_optimal_value_function_def}
    V^{*}_t(b) &= \min_{\pi} V^{\pi}_t(b)  \notag \\[0.5em]
    & = 
        \min \limits_{a \in \mathcal{A}} \left[ 
        \sum \limits_{s \in \mathcal{S}} C\left(s,a \right) \cdot b(s) + \gamma \cdot \sum \limits_{o \in \mathcal{O}} \mathrm{P}\left(o \vert b,a \right) \cdot V^{*}_{t+1}(b^a_o)
        \right],
\end{align}
\end{linenomath}
where $b^{a}_o$ is the belief that results from $b$ after executing action $a$ and observing $o$, and can be obtained from the POMDP model and Bayesian updating (e.g., demonstrated in \citep{braziunas2003pomdp}). Note that $\mathrm{P}(o|b,a)$ can be expressed as a function of the belief transition probability $\mathrm{P}(b_{o}^a|b,a)$, and the sum over $o$ can be transformed into a sum over $b$ (see \Cref{subsec:POMDP_reference_solution}).

One can also define an \emph{action-value function} $Q^{\pi}_t(b_t,a)$, which denotes the value of action $a$ at belief state $b$ under policy $\pi$ at time $t$ and continuing
optimally for the remaining timesteps until the end of the system lifetime \citep{braziunas2003pomdp}.
%
\begin{comment}
%
\begin{linenomath}
\begin{align}
\label{eq_q_function_def}
    Q^{{\pi}}_t(b,a) &= 
    \sum_{T=t}^{T_{\mathrm{end}}} \gamma^T * \mathbb{E}_{\pi} \left[ 
      C\left(b_T, \pi_T(b_T) \right) ~\bigr|~ b_t = b, ~a_t=a \right] \notag \\[0.5em]
    &= \sum \limits_{s \in \mathcal{S}} C\left(s,a \right)*b(s) + \gamma *\sum \limits_{o \in \mathcal{O}} \mathrm{P}\left(o \vert b,a \right)* V^{*}_{t+1}(b^{a}_o),
\end{align}
\end{linenomath}
%
\end{comment}
%
The optimal value function $V^{*}$ can be expressed as a minimization over the action-value function $Q$, and the optimal action-value function $Q^{*}$ satisfies the \emph{Bellman equation} \citep{braziunas2003pomdp, oliehoek2008optimal, kochenderfer2015decision}:
\begin{linenomath}
\begin{equation}
\label{eq_optimal_q_function_def}
   Q^{*}_t(b,a) = 
    \sum \limits_{s \in \mathcal{S}} C\left(s,a \right) \cdot b(s) + \gamma \cdot \sum \limits_{o \in \mathcal{O}} \mathrm{P}\left(o \vert b,a \right) \cdot \min_{a'\in\mathcal{A}} Q^{*}_{t+1}(b^{a'}_o,a'),
\end{equation}
\end{linenomath}
Lastly, the \emph{advantage function} $A^{\pi}_t(b,a)$ is a measure of the relative importance of each action \citep{wang2016dueling}:
\begin{linenomath}
\begin{equation}
\label{eq_definition_advantage}
    A^{\pi}_t(b,a) = Q^{\pi}_t(b,a) - V^{\pi}_t(b),
\end{equation}
\end{linenomath}
where the advantage of the optimal action $a^{*}$ is 0 \citep{wang2016dueling}:
\begin{linenomath}
\begin{equation}
\label{eq_advantage_of_optimal_action}
    Q^{\pi}_t(b,a^{*}) = V^{\pi}_t(b) \qquad \Longrightarrow \qquad
     A^{\pi}_t(b,a^{*}) = 0.
\end{equation}
\end{linenomath}

The optimal policy at every timestep can be easily extracted by performing a greedy selection over the optimal Q-value \citep{braziunas2003pomdp}:
\begin{linenomath}
\begin{equation}
\label{def_optimal_policy}
    \pi_t^{*}(b) = \argmin_{a \in \mathcal{A}} Q^{*}_t(b,a),
\end{equation}
\end{linenomath}
which is also the value- and advantage-minimizing action from \Cref{eq_optimal_value_function_def} and \Cref{eq_definition_advantage}, respectively.

The solution methods presented in \Cref{sec:NNs,sec:MCTS} have the goal of approximating $V$, $Q$, or $A$, from which the optimal policy can be extracted.
%
%
%
%
\begin{comment}
%
\begin{itemize}
    \item intro to POMDPs and reinforcement learning
    \item definition of (expected) LCC, V-, Q- and A-function
    \item time-dependent optimal policy
\end{itemize}
%
\end{comment}
%
%
%%%%%%%%%%%%%%%%%%%%%%%%%%%%%%%%%%%%%%%%%%%%%%
%%%%%%%%%%%%%%%%%%%%%%%%%%%%%%%%%%%%%%%%%%%%%%%
%
%
\subsection{POMDP reference solution}
\label{subsec:POMDP_reference_solution}
To evaluate the performance of approximate solutions, we also provide a reference solution for the POMDP model of \Cref{subsec:investigated_system}. It is computed with standard value iteration applied to a discretized belief MDP. The belief, in our case, is a vector comprising the posterior mean values of $D$ and $K$ from \Cref{Eq:mu_D_post,Eq:mu_K_post}:
\begin{linenomath}
\begin{equation}
\label{eq_belief_vect}
    \vect{b} = \begin{bmatrix}
          \mu_{D, t}'' \\
          \mu_{K, t}''
    \end{bmatrix}
\end{equation}
\end{linenomath}
The adapted version of \Cref{eq_optimal_value_function_def} for discretized beliefs is then \citep{cassandra1994acting,Noichl}: 
\begin{linenomath}
\begin{equation}
    \label{eq_V_opt_depending_on_b}
    V^{*}_t(\vect{b}) =  
    \min \limits_{a \in \mathcal{A}} \left[
    C\left(\vect{b}, a\right) + \gamma \cdot \sum \limits_{\vect{b}' \in \mathcal{B}} \mathrm{P}\left(\vect{b}' \vert \vect{b},a \right) \cdot V^{*}_{t+1}(\vect{b}'),
    \right]
\end{equation}
\end{linenomath}
where $C(\vect{b},a) = \sum \limits_{s \in \mathcal{S}} C(\vect{s},a) \cdot \vect{b}(\vect{s})$ (see \Cref{eq_optimal_value_function_def,eq_optimal_q_function_def}).

\begin{comment}
@@ Comparing with Eq. \Cref{eq_optimal_value_function_def}, here suddenly the cost is a function of belief and not the state. Also, the summation is over b' instead of o. I think this should be clarified @@
\end{comment}

Due to the linear Gaussian transition dynamics of this case study, the transition probabilities $\mathrm{P}\left(\vect{b}' \vert \vect{b},a \right)$ can be calculated analytically. The discretization of the belief and the computation of the probability tables is done according to \citep{straub2009stochastic}. \Cref{eq_V_opt_depending_on_b} is solved by backward induction for each discrete belief state. The resulting $\mean{\mathrm{LCC}}$ is verified by MCS. The discretization scheme is chosen such that 1) the value function of \Cref{eq_optimal_value_function_def} is estimated with a small error (compared to MCS on continuous belief space) and 2) such that the resulting policy is quasi-optimal (it performs better than every other solution found).

%
%%%%%%%%%%%%%%%%%%%%%%%%%%%%%%%%%%%%%%%%%%%%%%%%%%%%
%%%%%%%%%%%%%%%%%%%%%%%%%%%%%%%%%%%%%%%%%%%%%%%%%%%%
%%%%%%%%%%%%%%%%%%%%%%%%%%%%%%%%%%%%%%%%%%%%%%%%%%%%
%
%
%
%
%
%
\section{Neural networks}
\label{sec:NNs}
\subsection{Architecture}
\label{subsec:NN_architecture}
Our aim is an NN approach that is able to handle imperfect observations without the need for computing the belief. 
The NN needs to account for the time dependence of the value function for the finite horizon problem. This can be achieved by a network architecture that is able to handle sequential data, i.e., the observation-action history. For that, we adopt the basic structure of the action-specific deep recurrent Q-network \citep{hausknecht2015drqn}.

%(4d-vector) (single float)

The final NN architecture proposed in this work is depicted in \Cref{fig:snapshot_network_architecture}, which we name Action-specific Deep Dueling Recurrent Q-network (+RQN). At each timestep $t$, the two inputs of the network are the one-hot encoded action \citep{brownlee2020data} taken at $t-1$ and the scalar observation obtained at $t$. The outputs of the network are the estimated Q-values for each action at $t$.
The inputs are fed through two fully connected (FC) layers for feature extraction. The core of the network is formed by the Long Short-term Memory (LSTM) layer, which can resolve short as well as long-term dependencies through the hidden and cell states, respectively \citep{hochreiter1997lstm}. Depending on the observation-action history, these states take different values. Hence, the LSTM layer can be interpreted as a high-dimensional embedding of the history or a high-dimensional approximator of the belief state. The LSTM output is then fed through another FC layer for further feature extraction. To estimate the Q-values, the value and the advantage functions are first estimated separately and combined using a modified version of \Cref{eq_definition_advantage}, which is discussed in the section below. \citet{wang2016dueling} report that this configuration has superior performance compared to standard DQNs.

\nnarchitecture
%
%
%%%%%%%%%%%%%%%%%%%%%%%%%%%%%%%%%%%%%%%%%%%%%%%
%%%%%%%%%%%%%%%%%%%%%%%%%%%%%%%%%%%%%%%%%%%%%%%
%
%
\subsection{Q-values, loss, cost and weight updates}
%
%
%The straightforward way to compute the Q-values given the value function $V$ and advantage function $A$ is to use  \Cref{eq_definition_advantage}. However, the resulting equation suffers from a lack of identifiability: given a specific $Q$, we cannot recover $V$ and $A$ in a unique manner (adding and subtracting a constant yields the same $Q$) which leads to poor practical performance.

Instead of directly using \Cref{eq_definition_advantage}, \citet{wang2016dueling} propose to introduce the mean over the advantages as a correction term, which improves the stability of the optimization of the network parameters. Let $\vect{\theta}_t^j$ denote the parameters of all layers prior to the value-advantage split, $\vect{\upsilon}^j$ denote the parameters of the value stream, and $\vect{\alpha}^j$ the parameters of the advantage stream. The superscript $j=1,2,...,N_e$ refers to the weights at a certain iteration/epoch and hence highlights iterative convergence towards a set of weights that best approximate the true Q-value. Herein, an epoch consists of passing through the whole life cycle of a batch of sample trajectories, after which the weights get updated, and the next epoch starts. Since $\vect{\theta}$ does also include the hidden and cell states of the LSTM layer, it is dependent on the observation-action history, which is denoted with the subscript $t$. By contrast, $\vect{\alpha}$ and $\vect{\upsilon}$ stay constant for the whole life cycle (epoch).
The modified approximation for the Q-values is then \citep{wang2016dueling}:
\begin{linenomath}
\begin{multline}
\label{eq_modified_q_values}
    \mathtt{Q}_{t}^j(o_t,a~ |~a_{t-1}, \vect{\theta}_{t-1}^j,\vect{\alpha}^j, \vect{\upsilon}^j) = 
    \mathtt{V}_t^j(o_t~|~ a_{t-1},\vect{\theta}_{t-1}^j, \vect{\upsilon}^j)~ + \\ 
    \left(\mathtt{A}_t^j(o_t,a~|~a_{t-1}, \vect{\theta}_{t-1}^j,\vect{\alpha}^j)
    - \frac{1}{\abs{\mathcal{A}}} \sum_{a' \in \mathcal{A}} 
    \mathtt{A}_t^j(o_t,a' ~|~ a_{t-1}, \vect{\theta}_{t-1}^j,\vect{\alpha}^j) \right),
\end{multline}
\end{linenomath}
where $\mathtt{Q}_{t}^j(o_t,a~ |~ \vect{\theta}_{t-1}^j,\vect{\alpha}^j, \vect{\upsilon}^j)$ is the Q-value estimate for the action $a$ at time $t$ and epoch $j$ after observing $o_t$ and given the previous action $a_{t-1}$, the weights $\vect{\theta}_{t-1}^j$ (which embedded $o_{1:t-1}$ and $a_{1:t-2}$ through the hidden and cell states), $\vect{\alpha}^j$ and $\vect{\upsilon}^j$. Accordingly, $\mathtt{V}_t^j(o_t~|~ a_{t-1},\vect{\theta}_{t-1}^j, \vect{\upsilon}^j)$ does not use the weights of the separate advantage stream $\vect{\alpha}^j$ and vice versa.

%@@ suggest: replace $o$ with $h_t$ (history). Remove the subscript in the functions $Q$ etc. [check how we can make fig consistent.]
%
%Note that this modification alters the original meaning of the value and advantage functions, because they are now off by a constant.

%
\begin{comment}
@@ The network can estimate the Q-values at any point in time, for any observation-action history, given the network weights $\vect{\theta},\vect{\alpha},\vect{\upsilon}$.

The goal is to obtain the optimal sets of weights, which will result in the closest approximation of the Q-value.

If we assume that we know the optimal weights, we can compute the loss of our network as:
@@@
\end{comment}

To evaluate the performance of the network, i.e., the accuracy of the predicted Q-values, we need a target value for each pair of sample observation and action $o^{(i)}_t,~a^{(i)}_{t-1}$, which are passed as inputs to the network. To obtain a target value, we use the fact that the optimal Q-values follow the Bellman equation (\Cref{eq_optimal_q_function_def}). Therefore, we define the NN output $y_{\mathrm{NN,t}}^{(i),j}$ and the target value $y_{\mathrm{Tar,t}}^{(i),j}$ for a sample $(i)$ at a specific point in time $t$ and epoch $j$ as \citep{zhu2017action-specific}:
\begin{linenomath}
\begin{align}
\label{eq_dqn_nn_value}
y_{\mathrm{NN,t}}^{(i),j} &= %\min_{a \in \mathcal{A}}
\mathtt{Q}_{t}^j(o^{(i)}_t, a_{sel.} ~|~a_{t}, \vect{\theta}_{t-1}^j,\vect{\alpha}^j, \vect{\upsilon}^j)
\\[0.5em] 
y_{\mathrm{Tar,t}}^{(i),j} &= c^{(i)}_t + \gamma \cdot \min_{a' \in \mathcal{A}} \mathtt{Q}_{t+1}^j(o^{(i)}_{t+1}, a' ~|~ a_{t}, \vect{\theta}^{j,-}_t,\vect{\alpha}^{j,-},\vect{\upsilon}^{j,-}) \label{eq_dqn_target_value}
\end{align}
\end{linenomath}
where $a_{sel.}$ denotes the selected action under the behaviour policy ($\epsilon$-greedy, see \Cref{subsec:app_optimized_NN_parameters}) at $j$;
$c^{(i)}_t$ is the total cost sample at $t$ which includes the cost of a potential failure at $t$ and the cost of the latest selected action at $t-1$ under the behaviour policy at $j$. Moreover, $\mathtt{Q}_{t+1}^j(o^{(i)}_{t+1}, a' ~|~ \vect{\theta}^{j,-}_t,\vect{\alpha}^{j,-},\vect{\upsilon}^{j,-})$ denotes the Q-value estimate at $t+1$ and epoch $j$, after an action has been taken under the behaviour policy at $t$ and $j$ which, upon interaction with the environment, resulted in observation $o^{(i)}_{t+1}$. The "$-$" indicates that the parameters $\vect{\theta}^{j,-}_t,\vect{\alpha}^{j,-},\vect{\upsilon}^{j,-}$ belong to a separate \emph{target network} \citep{zhu2017action-specific}. More details on the sampling procedure and the target network are provided in \Cref{subsec:training_procedure}.
For training, we use the mean-squared error (MSE) \emph{loss function} %$\mathcal{L}_{\mathrm{MSE}}$ % which $y_{\mathrm{NN}}^{(i)}, y_\mathrm{Tar}^{(i)} \in \mathbb{R}$ takes the form of 
\citep{nielsen2015neural}: 

\begin{equation}
\label{eq_mse_loss}
    \mathcal{L}_{\mathrm{MSE}} \left(y_\mathrm{NN}, ~y_\mathrm{Tar}\right) 
    \coloneqq \left( y_{\mathrm{NN}} - y_{\mathrm{Tar}} \right)^2.
\end{equation}
%
\begin{comment}
\Cref{eq_mse_loss} measures the loss at a single timestep. Accumulating the loss over a whole life cycle trajectory sample  $\vect{y}_{\mathrm{NN}}^{(i)}, \vect{y}_\mathrm{Tar}^{(i)} \in \mathbb{R}^{T_{\mathrm{end}}-2}$ yields the LCC MSE loss $\mathcal{L}_{\mathrm{MSE}}^{\mathrm{LCC}}$:
%
\begin{linenomath}
\begin{equation}
\label{eq_mse_lcc_loss}
    \mathcal{L}_{\mathrm{MSE}}^{\mathrm{LCC}} \left(\vect{y}_\mathrm{NN}^{(i)}, ~\vect{y}_\mathrm{Tar}^{(i)}\right)
    %
    \coloneqq \sum_{t=1}^{T_{\mathrm{end}}-1} \left( y_{\mathrm{NN},t}^{(i)} - y_{\mathrm{Tar},t}^{(i)} \right)^2.
\end{equation}
\end{linenomath}
%
\end{comment}
For \emph{stochastic gradient descent}, a batch of $N_b$ samples is passed through the network to speed up training \citep{bottou1991stochastic}. 
On this basis, the MSE cost function accumulated over the whole life cycle is evaluated as
% The batch is used to calculate a more accurate representation of the networks performance, i.e. an average over the individual losses which is called the MSE \emph{cost function} $\mathcal{C}_{\mathrm{MSE}}$. For a batch size of $N_b$ it is calculated as \citep{nielsen2015neural}:
%
%\begin{equation}
%\label{eq_mse_cost}
%\mathcal{C}_{\mathrm{MSE}}\left(y_\mathrm{NN}^{(1:N_b)}, ~ y_\mathrm{Tar}^{(1:N_b)}\right) \coloneqq
%    \frac{1}{N_b}~ \sum_{i=1}^{N_b} 
%    \mathcal{L}_{\mathrm{MSE}} \left(\vect{y}_\mathrm{NN}^{(i)}, ~ \vect{y}_\mathrm{Tar}^{(i)}\right).
%\end{equation}
%
%\Cref{eq_mse_cost} measures the cost at a single timestep. Accumulating the loss over a whole life cycle trajectory sample  $\vect{y}_{\mathrm{NN}}^{(i)}, \vect{y}_\mathrm{Tar}^{(i)} \in \mathbb{R}^{T_{\mathrm{end}}-2}$ yields the LCC MSE cost $\mathcal{C}_{\mathrm{MSE}}^{\mathrm{LCC}}$:
%
\begin{linenomath}
\begin{equation}
\label{eq_mse_lcc_cost}
\mathcal{C}_{\mathrm{MSE}}^{\mathrm{LCC}}
\left(\vect{y}_\mathrm{NN}^{(1:N_b)}, ~
\vect{y}_\mathrm{Tar}^{(1:N_b)}\right) \coloneqq
    \frac{1}{N_b}~ 
    \sum_{t=1}^{T_{\mathrm{end}}-1}
    \sum_{i=1}^{N_b} 
    \mathcal{L}_{\mathrm{MSE}} \left(y_\mathrm{NN,t}^{(i)}, ~ y_\mathrm{Tar,t}^{(i)}\right) .
\end{equation}
\end{linenomath}

The NN weights are updated based on this cost function. The simplest gradient-based update scheme is \citep{nielsen2015neural,I2DL}:
\begin{linenomath}
\begin{equation}
\label{eq_sgd_weight_update}
    \vect{\upsilon}^{j+1} = \vect{\upsilon}^j - \eta \nabla_{\vect{\upsilon}}
    \mathcal{C}_{\mathrm{MSE}}^{\mathrm{LCC}}
    %\left(\vect{y}_\mathrm{NN}^{(1:N_b)},
    %~ \vect{y}_\mathrm{Tar}^{(1:N_b)}\right)
    ,
\end{equation}
\end{linenomath}
where $\eta$ is the learning rate. The weights $\vect{\alpha}^{j+1}$ and $\vect{\theta}^{j+1}$ are computed accordingly. Alternative update schemes such as RMSProp or Adam are available \citep[e.g.,][]{I2DL}.  
%
%
%%%%%%%%%%%%%%%%%%%%%%%%%%%%%%%%%%%%%%%%%%%%%%%
%%%%%%%%%%%%%%%%%%%%%%%%%%%%%%%%%%%%%%%%%%%%%%%
%
%
\subsection{Training procedure}
\label{subsec:training_procedure}
Each epoch $j$ is composed of a data collection and a training phase. The data collection part consists of simulating a batch of $N_b$ trajectories with the current network with weights $\vect{\theta}^j, \vect{\alpha}^j, \vect{\upsilon}^j$. We start by drawing initial samples $\vect{d}^{(i)}_0$ and $\vect{k}^{(i)}_0$ from their initial distributions and check for resulting failure costs $\vect{c}^{(i)}_{f,0}$. The actions $\vect{a}^{(i)}_0$ are fixed to $a_0$ (with action costs $\vect{c}^{(i)}_{a,0}=\vect{0}$), as observation-based action selection starts at $t=1$. Then, $\vect{d}^{(i)}_0$, $\vect{k}^{(i)}_0$, $\vect{a}^{(i)}_0$ are passed to the environment which returns $\vect{d}^{(i)}_1$, $\vect{k}^{(i)}_1$ and $\vect{c}^{(i)}_{f,1}$ according to the dynamics in \Cref{tab:action_effect_on_samples_and_belief}. For $t=1,...,T_{\mathrm{end}-1}$, observations $\vect{o}^{(i)}_t$ are generated from $\mathcal{N}(\vect{d}^{(i)}_t, \sigma_E)$ and passed with $\vect{a}^{(i)}_{t-1}$ to the network which outputs the Q-values. The behaviour policy at epoch $j$ selects the next action according to the $\epsilon$-greedy scheme, where a random action is selected with probability $\epsilon$ (for exploration) and the action with minimal Q-value is selected with probability $1-\epsilon$ (for exploitation). The chosen action $\vect{a}^{(i)}_t$ together with $\vect{d}^{(i)}_t$ and $\vect{k}^{(i)}_t$ is passed to the environment that simulates the system for one timestep and returns $\vect{d}^{(i)}_{t+1}$, $\vect{k}^{(i)}_{t+1}$ and $\vect{c}^{(i)}_{f, t+1}$. This alternating interaction between network and environment continues until the end of the system lifetime is reached. The samples $\vect{o}^{(i)}_{1:T_{\mathrm{end}-1}}$, $\vect{a}^{(i)}_{0:T_{\mathrm{end}-2}}$ and $\vect{c}^{(i)}_{0:T_{\mathrm{end}}} = \vect{c}^{(i)}_{f, 0:T_{\mathrm{end}}} +  \vect{c}^{(i)}_{a, 1:T_{\mathrm{end}-1}}$ are then stored for the training phase.

Once a batch of sample trajectories has been collected, the training phase starts. Herein, the batch is again fed through the network sequentially, and the cost is accumulated over the whole life cycle. For the computation of the individual MSE loss terms, a target network is defined such that the values of the target network weights are clones of the original network weights: $\theta^{j,-} = \theta^j$, $\alpha^{j,-} = \alpha^j$, $\upsilon^{j,-} = \upsilon^j$. At each time $t$, $\vect{o}^{(i)}_t$ and $\vect{a}^{(i)}_{t-1}$ are the inputs of the network; $\vect{o}^{(i)}_{t+1}$ and $\vect{a}^{(i)}_t$ are the inputs of the target network. The target NN outputs are greedily selected over the respective Q-values (as opposed to the $\epsilon-$greedy behaviour policy used for trajectory sampling, hence this is \emph{off-policy} learning \citep{mctsandrf}) according to \Cref{eq_dqn_nn_value,eq_dqn_target_value}. The batch cost at $t$ is computed with a batch-averaged version of \Cref{eq_mse_loss} and added to the total cumulative cost. This process continues until the end of the life cycle is reached, and the LCC MSE cost has been computed according to \Cref{eq_mse_lcc_cost}. Then, the LSTM is unrolled, the loss is backpropagated through time \citep{hochreiter1997lstm} and the weights are adjusted according to the chosen update scheme (e.g., \Cref{eq_sgd_weight_update}). After updating, the learning procedure continues with the next epoch until the weights have converged. The weights of the target network are updated periodically every $p$ epochs to ensure stable optimization \citep{mnih2015human}.

The hyperparameter tuning procedure, either by grid search or by some heuristics, is outlined in \Cref{sec:app_nn_specs}. 

\begin{comment}
%
\begin{enumerate}
\label{enum_steps_at_each_timestep}
    \item \label{item_1} Check for failure ($D_t>d_{cr}$) and pay the potential failure cost $c_{F_t}$
    \item \label{item_2} Draw a measurement sample $O_t \sim \mathcal{N}(D_t, \sigma_E)$
    \item \label{item_3} Decide for an action $A_t$ and pay the corresponding action cost $c_{A_t}$
    \item \label{item_4} Set $D_{t+1}$ and $K_{t+1}$ according to the chosen action
    \item \label{item_5} Repeat
\end{enumerate}
%
%
%
\begin{itemize}
    \item we propose +RQN architecture (instead of ADDRQN ) which stands for Action-specific Dueling Deep Recurrent Q-Network with reference to two papers
    \item Q-value computed from A- and V-values (dueling approach)
    \item LSTM used to deal with time-dependence of optimal solution
    \item inclusion of target network
    \item loss function as deviation from Bellman equation
    \item whole life cycles fed through sequentially before backpropagation
    \item architecture figure and NN specifications in 
    \item optimization via grid search
\end{itemize}
%
%
\end{comment}

%
%
%
%%%%%%%%%%%%%%%%%%%%%%%%%%%%%%%%%%%%%%%%%%%%%%%%%%%%
%%%%%%%%%%%%%%%%%%%%%%%%%%%%%%%%%%%%%%%%%%%%%%%%%%%%
%%%%%%%%%%%%%%%%%%%%%%%%%%%%%%%%%%%%%%%%%%%%%%%%%%%%
%
%
%
%
\section{MCTS}
\label{sec:MCTS}
\subsection{Functionality}
\label{subsec:functionality}
\emph{Monte Carlo tree search} (MCTS) arises from the combination of tree search and Monte Carlo sampling \citep{metropolis1949mcmethod}. Classically, games have been modeled with game trees, where the root is the starting position, leaves are possible ending positions, and each edge represents a possible move \citep{tarsi1983trees}. To select the best action at a given node (position), one needs to know its consequences. Small games can be solved by constructing the full game tree and using backwards induction \citep{gibbons1992primergametheory}. However, for more complex games (e.g., chess, Go), this is practically impossible. Hence, one needs an estimator of the preference for each resulting position. Defining the value of each node as an \emph{expected outcome} given random play opened the door for the use of Monte Carlo, which specifies node values as random variables and characterizes game trees as probabilistic  \citep{abramson2014expectedoutcomemodel}. In \citep{silver2010pomcp}, MCTS was extended to partially observable environments.

The MCTS algorithm consists of four main steps: selection, expansion, rollout, and backpropagation. In the selection step, the algorithm traverses the tree from the root to a leaf node using a selection policy (see \Cref{subsec:uct_as_behavioural_policy}). In the expansion step, the algorithm adds a child node to the selected leaf node. In the rollout step, the algorithm performs a simulation from the newly added node until the end of the lifetime by choosing uniformly random actions, i.e., $p(a_i)=\frac{1}{4}$. In the backpropagation step, the algorithm updates the statistics of all nodes along the path from the selected node to the root node based on the simulation outcome \citep{browne2012survey}. The Q-value of an action $a$ for a given observation-action history $h$ at time $t$ is the updated statistic at an action and is computed as:
%
\begin{comment}
The \emph{rollout} phase denotes a simulation from a leaf node (which denotes some state of the system) until the end of the lifetime by choosing random actions, i.e., $p(a_i)=\frac{1}{4}$. During the rollout, all incurring costs (action and failure costs) are discounted, summed and returned as a sample remaining LCC under a random action policy. MCTS employs Monte Carlo sampling to estimate the Q-value of an action $a$ for a given observation-action history $h$ at time $t$ as:
\end{comment}
%
%
\begin{linenomath}
\begin{equation}
\label{eq_MCTS_Q_update}
Q_t(h,a) \approx \frac{1}{N(h,a)} \sum_{i=1}^{N(h,a)} q^{(i)}_t(h,a),
\end{equation}
\end{linenomath}
where $N(h,a)$ is the total number of samples used for the estimation, or the current visitation counter of the respective action node, and $q^{(i)}_t(h,a)$ are the individual (backpropagated) results at time $t$.
%
%
%%%%%%%%%%%%%%%%%%%%%%%%%%%%%%%%%%%%%%%%%%%
%%%%%%%%%%%%%%%%%%%%%%%%%%%%%%%%%%%%%%%%%%%
%
%
\subsection{UCT for action selection}
\label{subsec:uct_as_behavioural_policy}
%

%@@ This section is a bit too compact, I find it difficult to understand how this works. Maybe we can go together through this. 

To make use of the exploitation-exploration tradeoff  \citep{dong2020deep}, we implement the Upper Confidence Bound for Trees (UCT) algorithm. The UCT selects the next action $A_t$ based on minimizing the estimation of the Q-value for each action (exploitation) minus an exploration term \citep{kocsis2006uct}:
\begin{linenomath}
\begin{equation}
\label{eq_uct}
A_t = \argmin_{a \in \mathcal{A}} \text{UCT}_t(h,a) = \argmin_{a \in \mathcal{A}} Q_t(h,a) - c \cdot \sqrt{\frac{\ln N(h)}{N(h,a)}},
\end{equation}
\end{linenomath}
where $c$ is an adjustable constant that enables a trade-off between exploration and exploitation, and $N(h)$ is the visitation counter of the parent node such that $N(h) = \sum_a N(h,a)$. A pseudocode for the implementation of MCTS for POMDPs is given in \citep{silver2010pomcp}.

Each simulation starts by sampling an initial state from the current belief state, which is, for our case study, described in \Cref{eq_current_belief_state_for_sampling}:
\begin{linenomath}
\begin{equation}
\label{eq_current_belief_state_for_sampling}
    \begin{bmatrix}
        D_t \\
        K_t
    \end{bmatrix}
    \sim \mathcal{N} \left(
    \begin{bmatrix}
          \mu_{D, t}'' \\
          \mu_{K, t}''
    \end{bmatrix},~
    \begin{bmatrix}
          \sigma_{D, t}^{\prime \prime 2} &
          \rho_t'' \sigma_{D, t}'' \sigma_{K, t}''\\
          \rho_t'' \sigma_{D, t}'' \sigma_{K, t}'' & \sigma_{K, t}^{\prime \prime 2}
    \end{bmatrix}
    \right).
\end{equation}
\end{linenomath}
\citet{silver2010pomcp} propose a samples-based approximation of the belief state for the general case when the belief state is not analytically available. We have not implemented this in the case study, hence one should keep in mind that an MCTS without the belief is likely to perform worse.

\begin{comment}
each node contains also a set of particles $B(h)$. For every history $h$ encountered during the simulation, the simulation state is added to the belief state $B(h)$. Each simulation starts by sampling the current belief state
from a starting state that is sampled from the current belief state $B(h_t)$. We can calculate the belief state analytically for our case study, hence we use the exact belief state instead of the collection of simulation states.
\end{comment}

The tuning of the MCTS parameters is outlined in \Cref{sec:app_MCTS_optimization}.

\begin{comment}
%
should be much shorter than section for NNs
\begin{itemize}
    \item concept of MCTS (value of state as expected reward given random play, every step identical, arbitrary starting point possible)
    \item use of UCT
    \item analytical beliefs used for sampling instead of visited states (difference to Silver \& Veness paper)
    \item important parameters (discretization of obs., exploration constant, tree iterations, rollout runs)
    \item (optimization technique?)
\end{itemize}
%
\end{comment}
%
%
%
%%%%%%%%%%%%%%%%%%%%%%%%%%%%%%%%%%%%%%%%%%%%%%%%%%%%
%%%%%%%%%%%%%%%%%%%%%%%%%%%%%%%%%%%%%%%%%%%%%%%%%%%%
%%%%%%%%%%%%%%%%%%%%%%%%%%%%%%%%%%%%%%%%%%%%%%%%%%%%
%
%
%
%
\section{Results}
\label{sec:Results}
\subsection{Metrics for comparison}
\label{sec:Metric_for_comparison}
We employ several metrics to assess the performance of the NN and the MCTS approaches and to compare the results to the POMDP reference solution. 

Firstly, we evaluate the computation time needed by the methods, including training and testing times. 

Secondly, their computational performance is compared through the LCC's expected value and the standard deviation for the identified policies. Thereby, $\mean{\mathrm{LCC}}$ is approximated with Monte Carlo (MC) samples for both methods. The optimal solution curve obtained by evaluating the POMDP with VI (see \Cref{subsec:POMDP_reference_solution}) serves as a reference. We additionally provide the performance of a benchmark policy that consists of choosing action $a_1$ in every timestep, irrespective of the observation.

Thirdly, we investigate the policies obtained from each method. The analysis comprises a statistical representation of the actions taken at each timestep to reveal potential tendencies, as well as a depiction in the belief space for policy extraction.

%
%
%
%
%%%%%%%%%%%%%%%%%%%%%%%%%%%%%%%%%%%%%%%%%%%%%%%%%%%%
%%%%%%%%%%%%%%%%%%%%%%%%%%%%%%%%%%%%%%%%%%%%%%%%%%%%
%%%%%%%%%%%%%%%%%%%%%%%%%%%%%%%%%%%%%%%%%%%%%%%%%%%%
%
%
%
%
%
%

%
\subsection{Computation time}
\label{subsec:computation_time}
All computations are performed on a Fujitsu Celcius R970 PC comprising an NVIDIA GP104GL (Quadro P4000) 8118 MB GPU and Intel Xeon Silver 4114 2.20 GHz: 10 Cores 20 Logical Processor. To accelerate the computation, training and testing of the NNs is conducted on the GPU, whereas MCTS is implemented with CPU parallelization. 

With these specifications, the process of training and testing a single NN took 45 seconds (25 seconds of training and 20 seconds of testing $10^6$ sample trajectories). In training, we consider different hyperparameter configurations following Section \ref{subsec:app_optimized_NN_parameters}, which leads to a total training time of approx. $150$min.

By contrast, with MCTS there is no distinct training phase. Nevertheless, it is necessary to find good MCTS parameters, as described in Section \ref{subsec:tunable_MCTS_parameters}. This is a time-consuming process, because testing is expensive with MCTS. With the chosen parameter setting, generating 1000 trajectories for testing takes 20 minutes. For this reason, NN training is ultimately significantly cheaper and more straightforward.

Once the NN is trained or the MCTS setting is fixed, evaluating the policy is efficient. For NN, the computational time is negligible; for MCTS, it is in the order of seconds. 
%
%%%%%%%%%%%%%%%%%%%%%%%%%%%%%%%%%%%%%%%%%%%%%%%%%%%%%%5
%%%%%%%%%%%%%%%%%%%%%%%%%%%%%%%%%%%%%%%%%%%%%%%%%%%%%%
%
\subsection{Performance}
\label{subsec:numerical_performance}
\begin{figure}[t]
    \centering
    \includegraphics[width=0.95\textwidth]{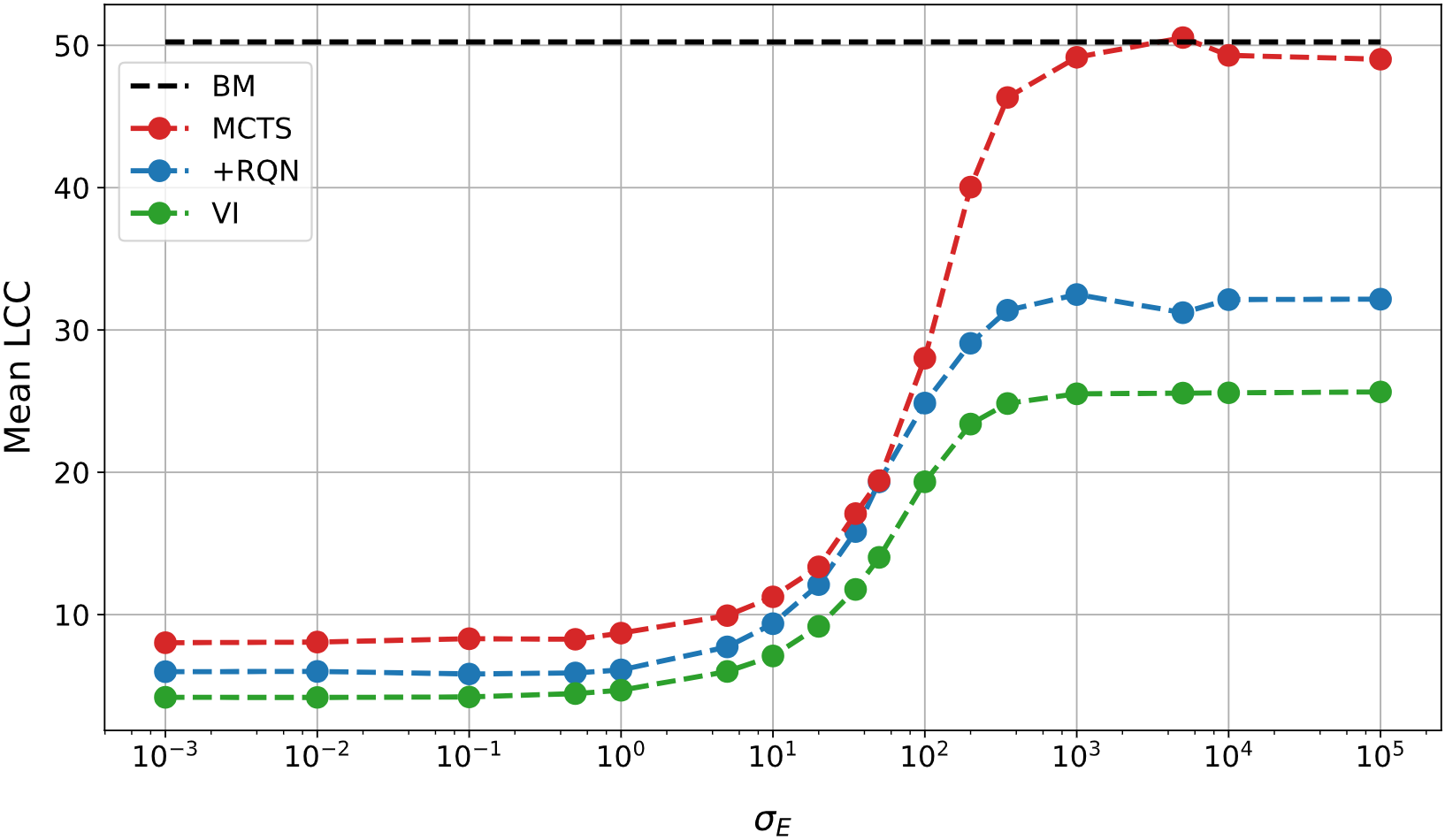}
    %\captionsetup{width=0.95\textwidth}
    \caption{Comparison of achieved mean LCC of +RQN (blue), MCTS (red), value iteration (green), and our a priori benchmark (black) for different measurement errors, where the policies of the +RQN, MCT and VI are averaged over $10^6$, $2\times10^3$ and $2\times10^6$ trajectories, respectively.}
    \label{fig:measurement_error_evolution}
\end{figure}

\Cref{fig:measurement_error_evolution} shows the mean LCC achieved by the +RQN, MCTS, VI, and the basic benchmark in function of the observation error. Firstly, all curves have a characteristic shape which consists of two saturation regions $\sigma_E<0.5$ (essentially corresponding to perfect observations) and $\sigma_E>10^3$ (uninformative observations) and a smooth transition in-between. Both the NN and MCTS methods perform worse than the optimal solution. 
%, wherein the difference increases with increasing measurement error - this is true when looking at the absolute value, but not the relative value
However, the NN consistently outperforms the MCTS method, which performs especially poorly under high observation errors.   

\begin{figure}[H]
    \centering
    \includegraphics[width=0.95\textwidth]{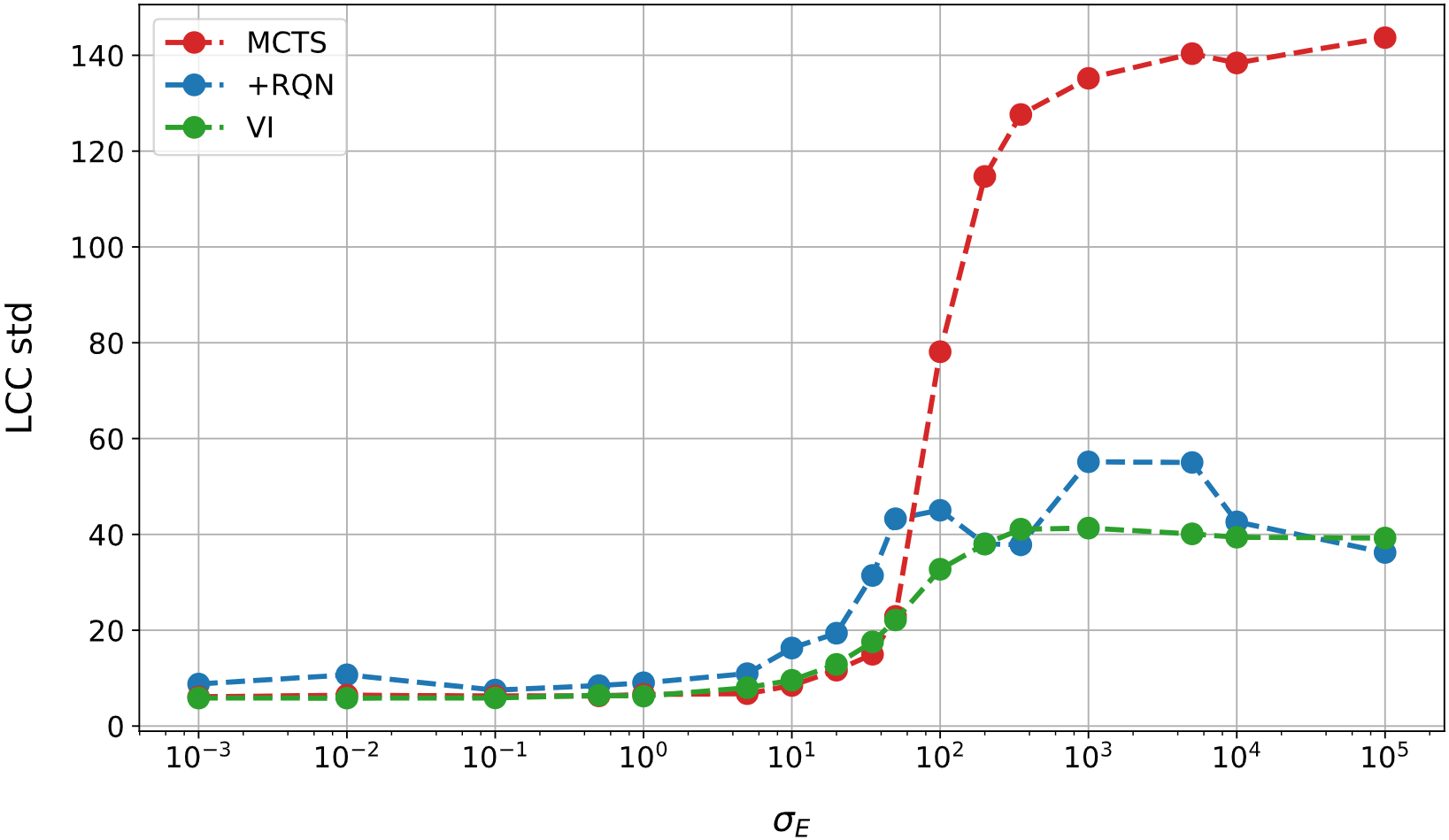}
    %\captionsetup{width=0.95\textwidth}
    \caption{Comparison of achieved LCC standard deviation of +RQN (blue), MCTS (red), value iteration (green) for different measurement errors, where the policies of the +RQN, MCT and VI are averaged over $10^6$, $2\times10^3$ and $2\times10^6$ trajectories, respectively.}
    \label{fig:measurement_error_evolution_std}
\end{figure}

\Cref{fig:measurement_error_evolution_std} shows the standard deviation of the resulting LCC in function of the observation error. The standard deviation increases with increasing observation error, which is to be expected. The NN generally leads to a slightly higher LCC standard deviation than the VI reference solution, although with some exceptions. By contrast, the MCTS results in a low LCC standard deviation for small $\sigma_E$ and in a very large one for large $\sigma_E$.

%There are similarities to \Cref{fig:measurement_error_evolution}, namely that the curves loosely have a similar shape and that the VI solution generally provides a lower bound - although with some noticeable exceptions. Nevertheless, there are also key differences: the NN curve is not smooth, especially for high observation errors, where it seemingly increases or decreases in a random fashion. Moreover, for $\sigma_E<100$ the MCTS performance is on par or even better than the VI solution, but for $\sigma_E\geq100$ its performance decreases significantly.
%
%
%
%
%
%%%%%%%%%%%%%%%%%%%%%%%%%%%%%%%%%%%%%%%%%%%%%%%%%%%%%%5
%%%%%%%%%%%%%%%%%%%%%%%%%%%%%%%%%%%%%%%%%%%%%%%%%%%%%%
%
\subsection{Policy comparison}
\label{subsec:policy_comparison}
\begin{figure}[H]
    \centering
    \captionsetup[subfigure]{position=t, justification=centering, captionskip=-2pt}
     \subfloat[VI: $\sigma_E=0.5$]{
    \includegraphics[width=0.460\textwidth]{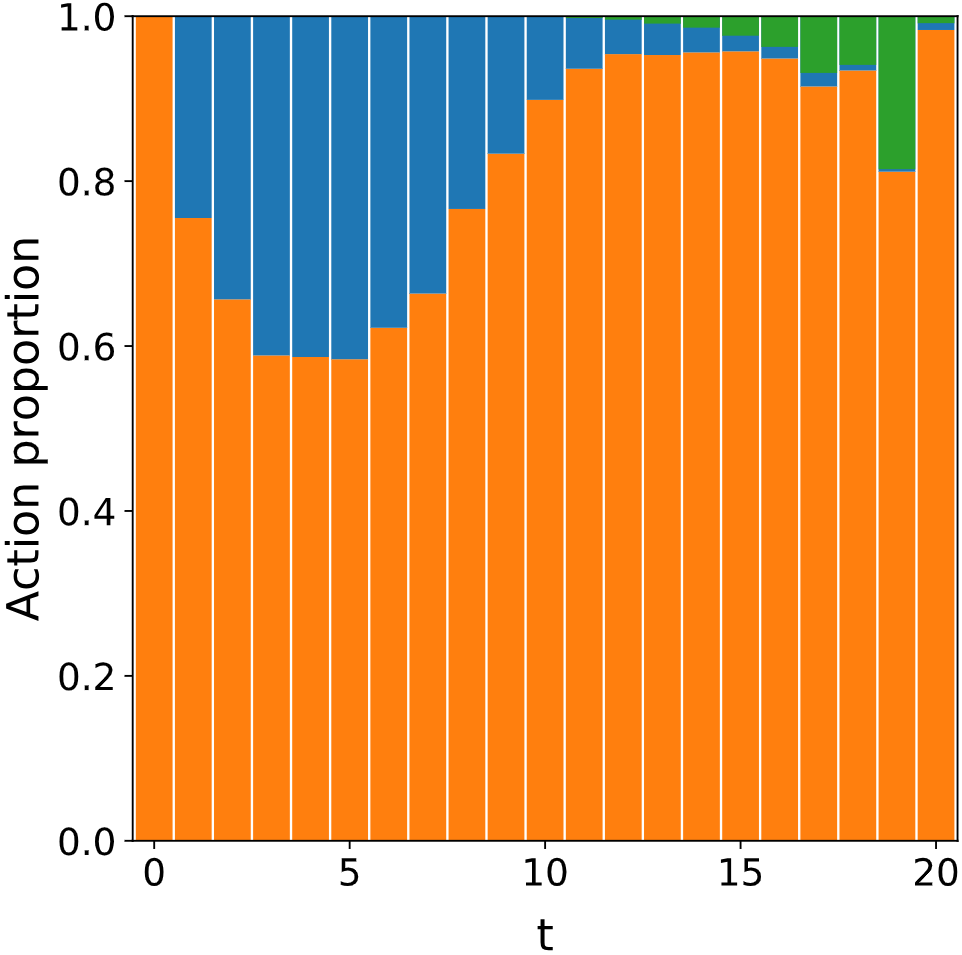}
    }
    \subfloat[VI: $\sigma_E=50$]{
    \includegraphics[width=0.54\textwidth]{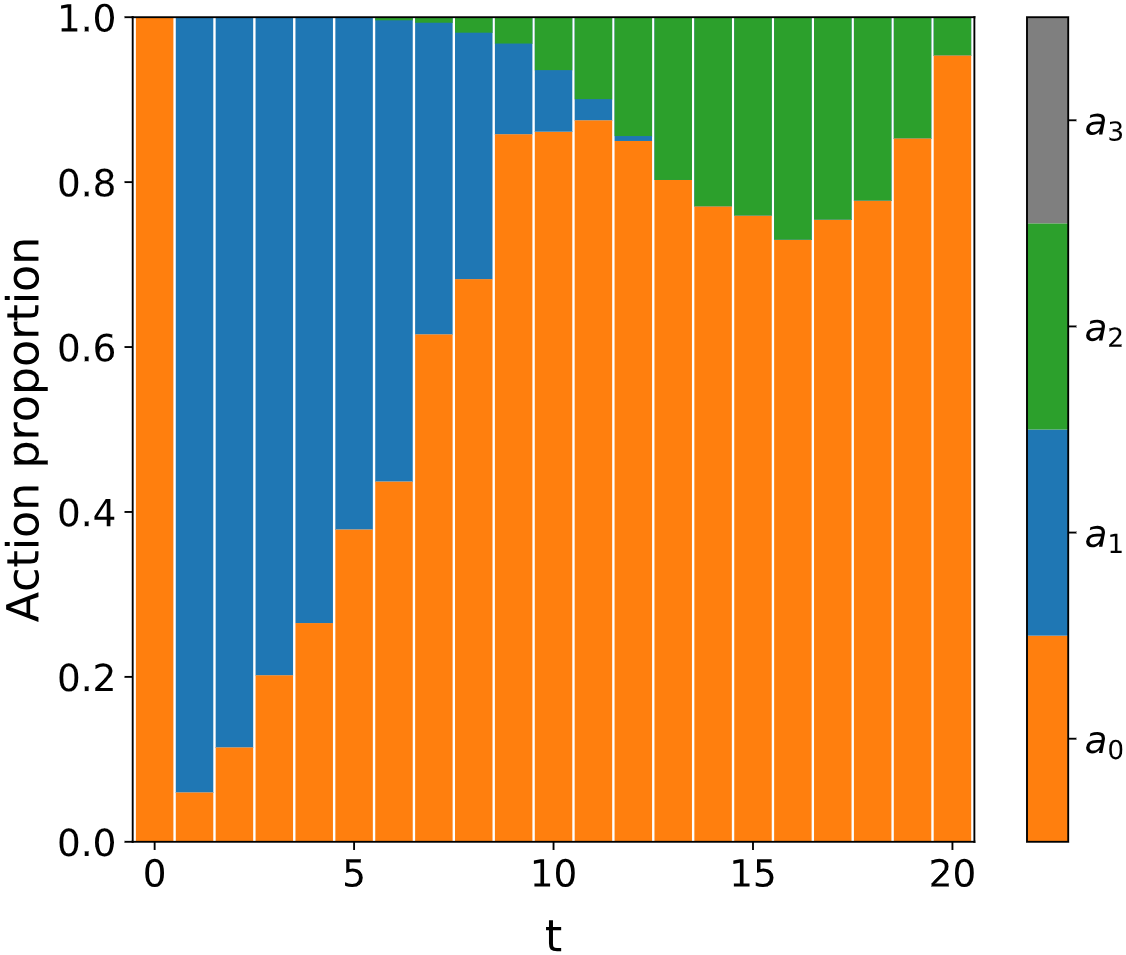}
    }
    \vspace{-0.2cm}
    \newline
    %\label{fig:action_statistics_comparison}
%\end{figure}
%\begin{figure}[H]
    %\setcounter{subfigure}{2}
    %\centering
    %\captionsetup[subfigure]{position=t, justification=centering, captionskip=-2pt}
    \subfloat[+RQN: $\sigma_E=0.5$]{
    \includegraphics[width=0.46\textwidth]{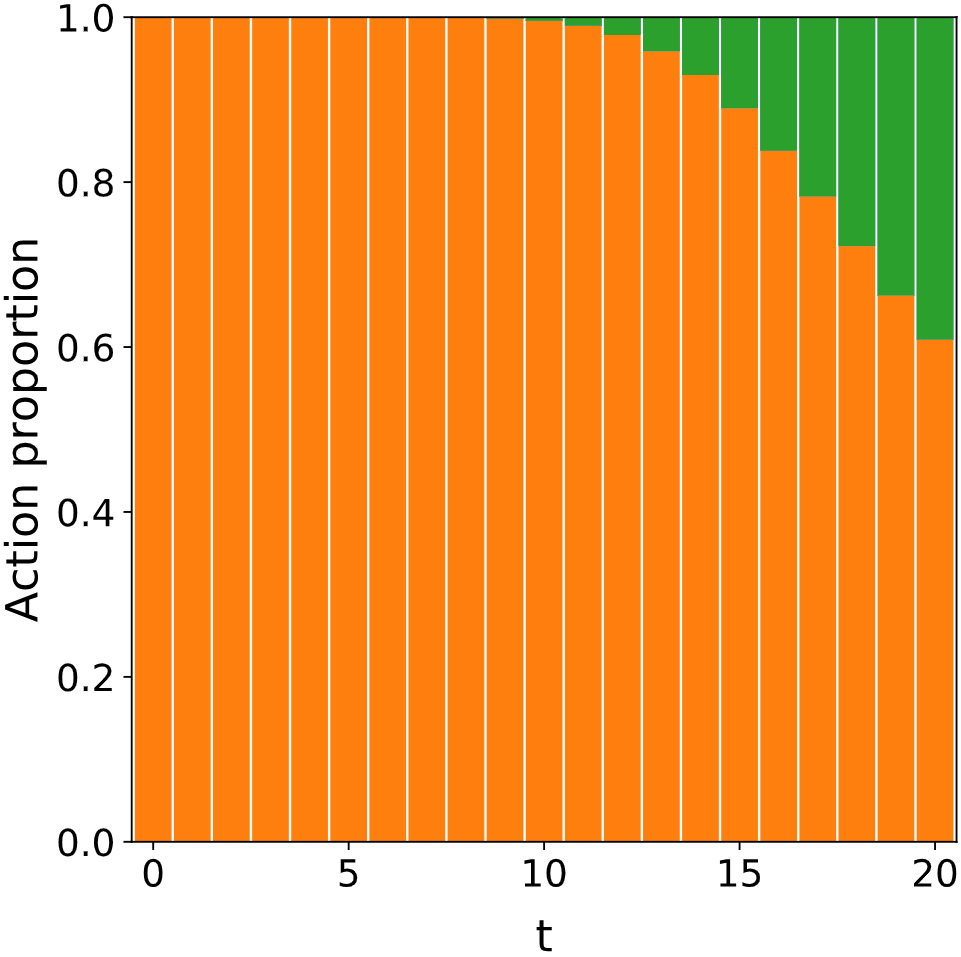}
    }
    \subfloat[+RQN: $\sigma_E=50$]{
    \includegraphics[width=0.54\textwidth]{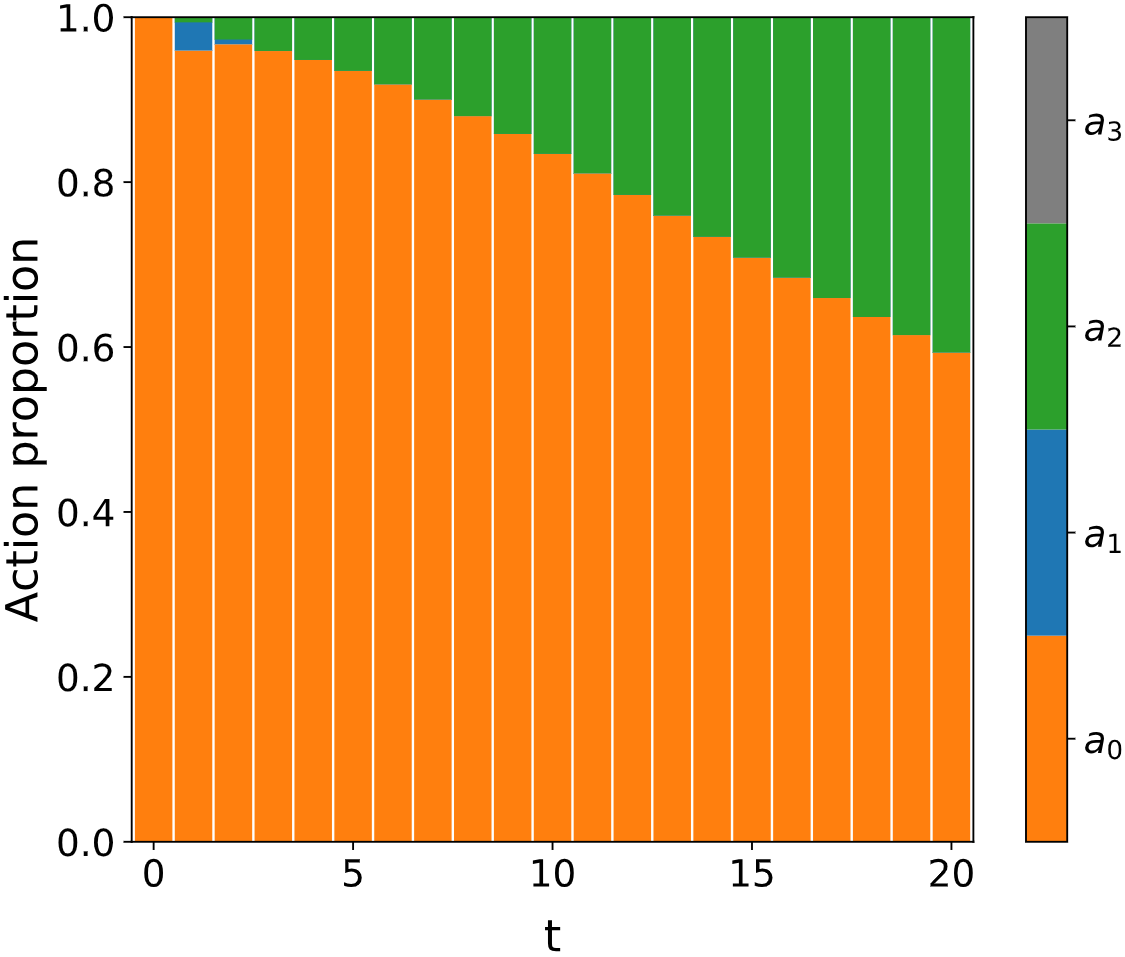}
    }
%\end{figure}
%\begin{figure}[H]
%    \setcounter{subfigure}{4}
%    \centering
%    \captionsetup[subfigure]{position=t, justification=centering, captionskip=-2pt}
\vspace{-0.2cm}
    \newline
    \subfloat[MCTS: $\sigma_E=0.5$]{
    \includegraphics[width=0.46\textwidth]{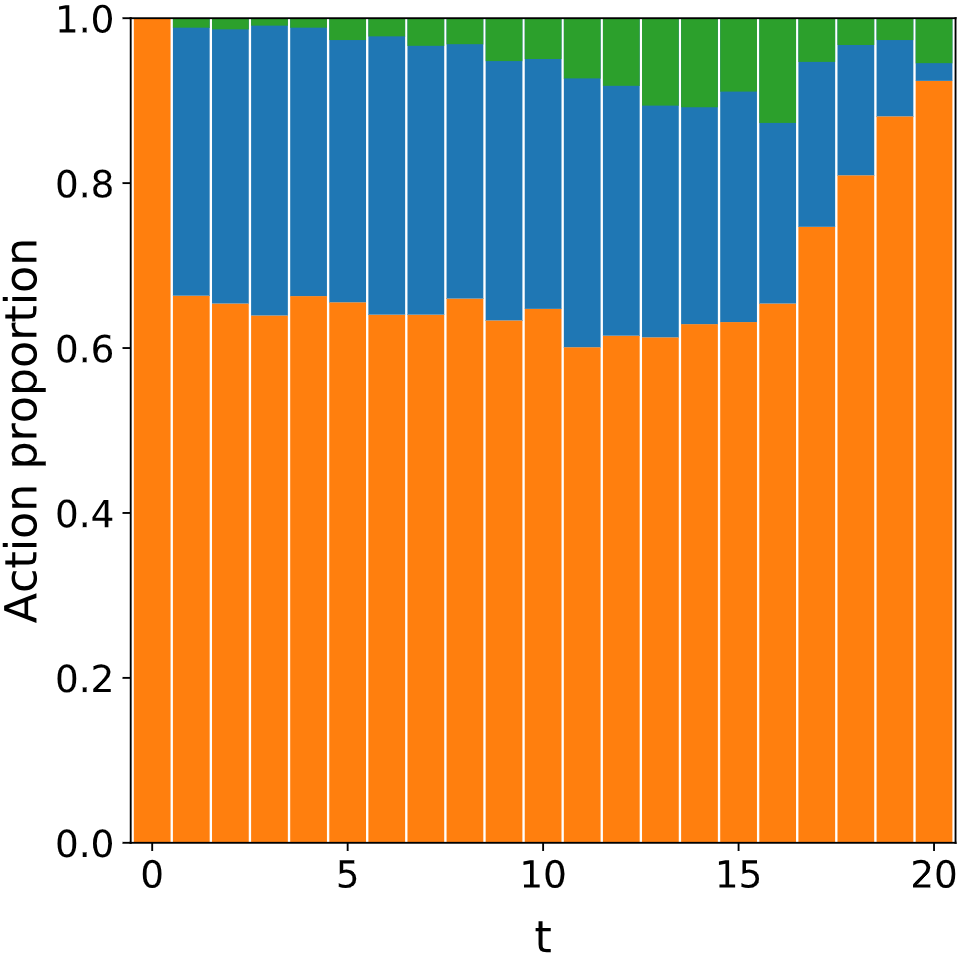}
    }
    \subfloat[MCTS: $\sigma_E=50$]{
    \includegraphics[width=0.54\textwidth]{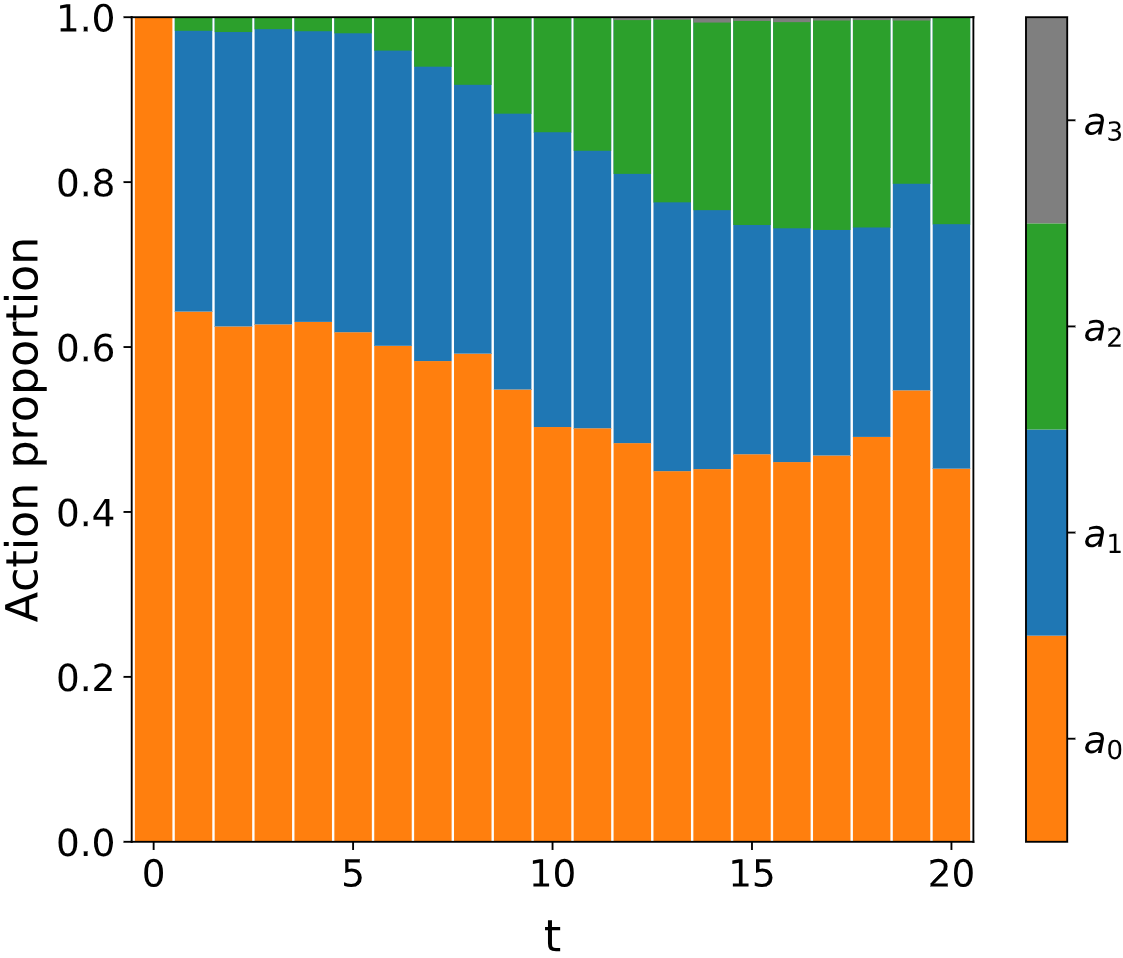}
    }
    %\captionsetup{width=0.95\textwidth}
    \caption{Temporal evolution of the action statistics represented by bar charts for VI (a) \& (b), +RQN (c) \& (d), and MCTS (e) \& (f) generated with $2\times10^6$, $10^6$, $2\times10^3$ and MC samples, respectively. }
    \label{fig:action_statistics_comparison}
\end{figure}

\Cref{fig:action_statistics_comparison} depicts the identified strategy profiles for the +RQN, MCTS, and VI in a statistical sense for the selected cases of $\sigma_E=\{0.5,~50\}$. 

The reference VI method utilizes mainly $a_1$ in the first half of the system lifetime and employs $a_2$ in the second half. More maintenance is performed when the observation error is larger; for $\sigma_E=50$, action $a_1$ is implemented early on in all cases, i.e., independent of the observation. Action $a_3$ is avoided, presumably due to its large cost.

The actions selected by the NN, as shown in panels (c) and (d), differ significantly from those of the reference solution. 
Note that the policies obtained with the NN vary substantially among repeated training runs, even if they lead to similar $\mean{\mathrm{LCC}}$. The results in \Cref{fig:action_statistics_comparison} correspond to a single trained 
NN for each observation error; with other trained NN instances, different proportions of $a_0,a_1,a_2$ are observed. 
%These differences can be quite pronounced, e.g., some runs result in no use of $a_0$ at all, while others result in a larger proportion of $a_1$ with reduced $a_2$.
In all trained NN, we observe that for $\sigma_E<200$, the NN employs solely $a_2$ for failure prevention; $a_1$ is involved only for higher observation errors.  

By contrast, MCTS has a similar strategy profile over all observation errors: about 30\% use $a_1$ at every timestep. The only difference observed for larger observation errors is the increased use of $a_2$ in the second half of the system lifetime with increasing measurement errors.
Interestingly, for small $\sigma_E$, the statistic of the selected actions with MCTS is closer to the reference solution than the one of NN, even if the expected LCC achieved with the NN is smaller than the one achieved with MCTS.

\begin{figure}[H]
\setcounter{subfigure}{0}
    \centering
    \captionsetup[subfigure]{position=t, justification=centering}
    \subfloat[t=1]{
    \includegraphics[width=0.46\textwidth]{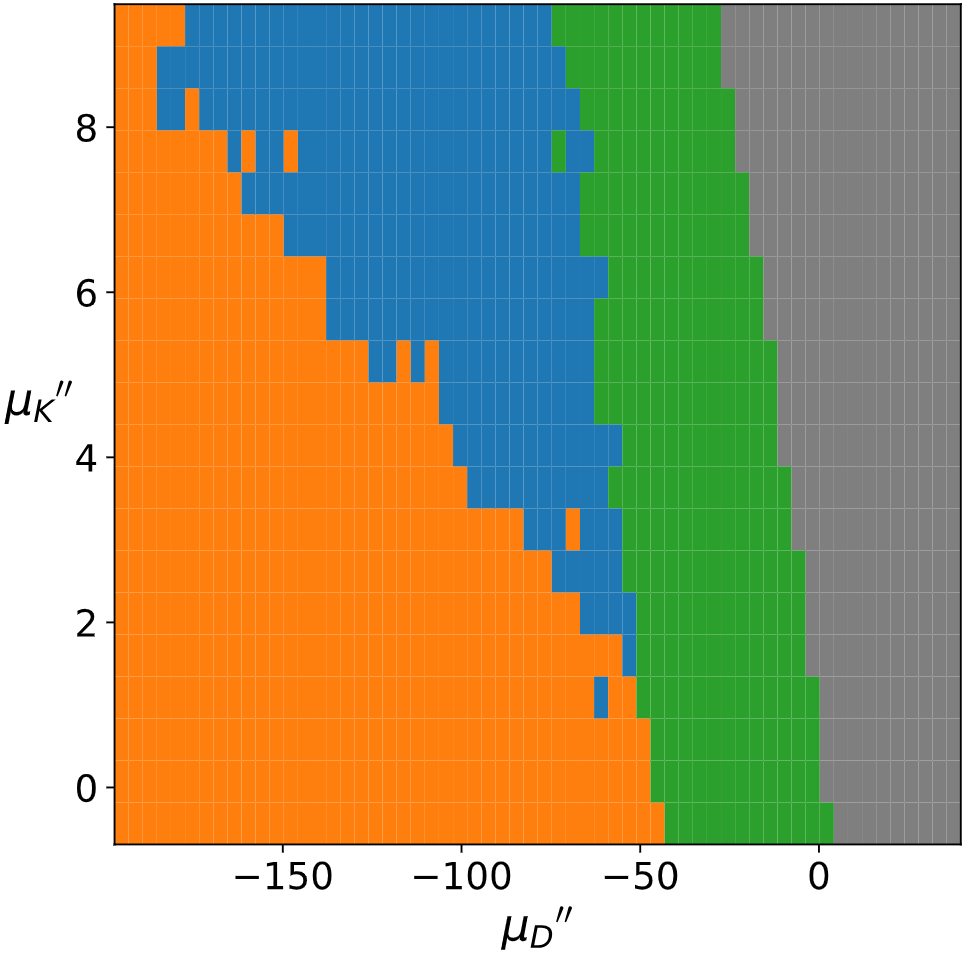}
    }
    \subfloat[t=10]{
    \includegraphics[width=0.54\textwidth]{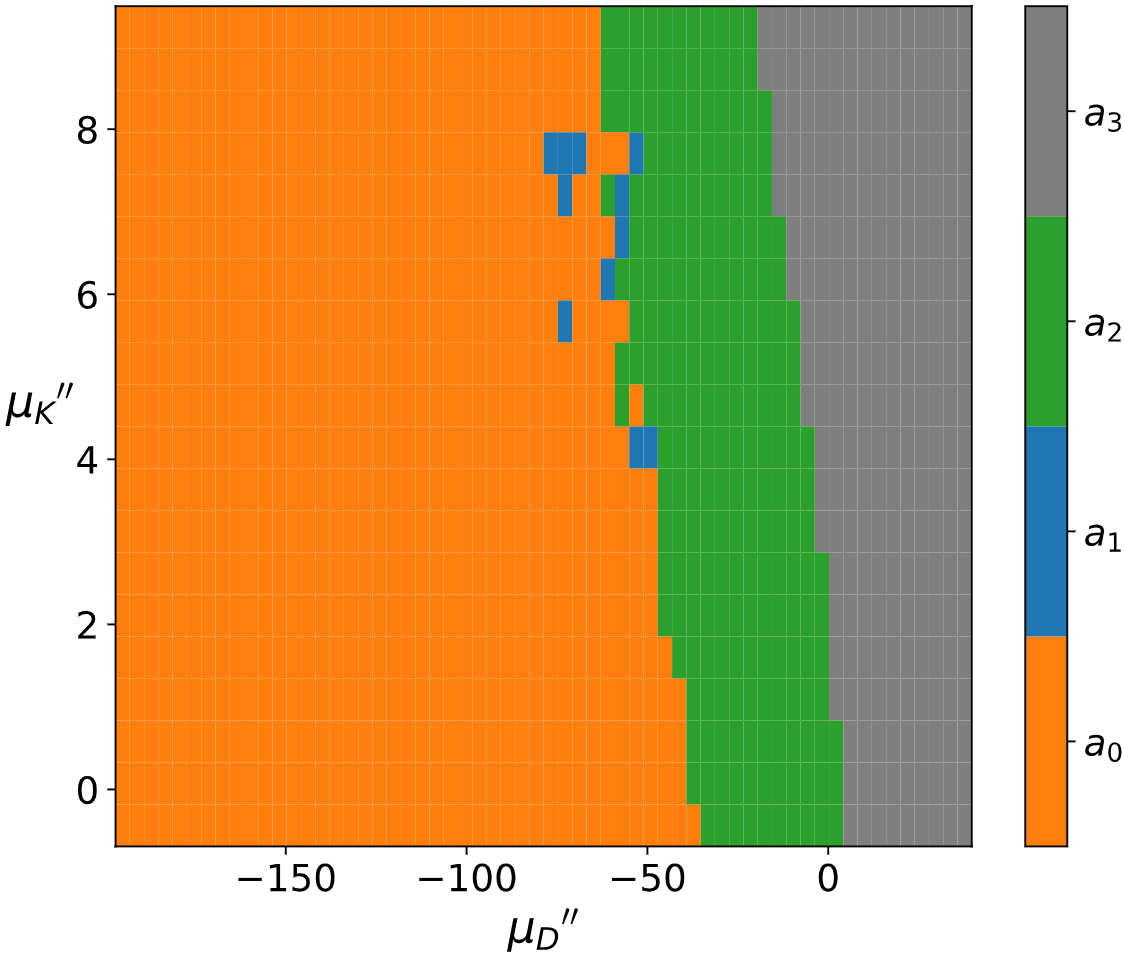}
    }
    \newline
    \subfloat[t=18]{
    \includegraphics[width=0.46\textwidth]{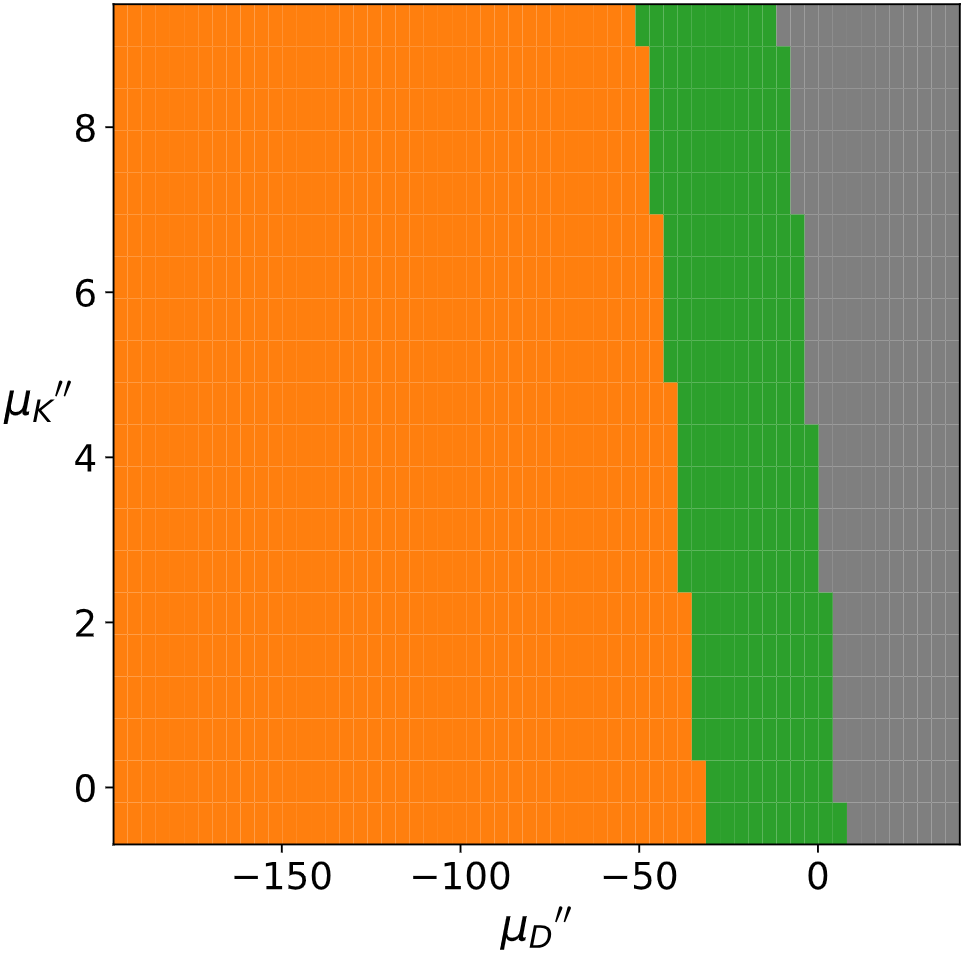}
    }
    \subfloat[t=20]{
    \includegraphics[width=0.54\textwidth]{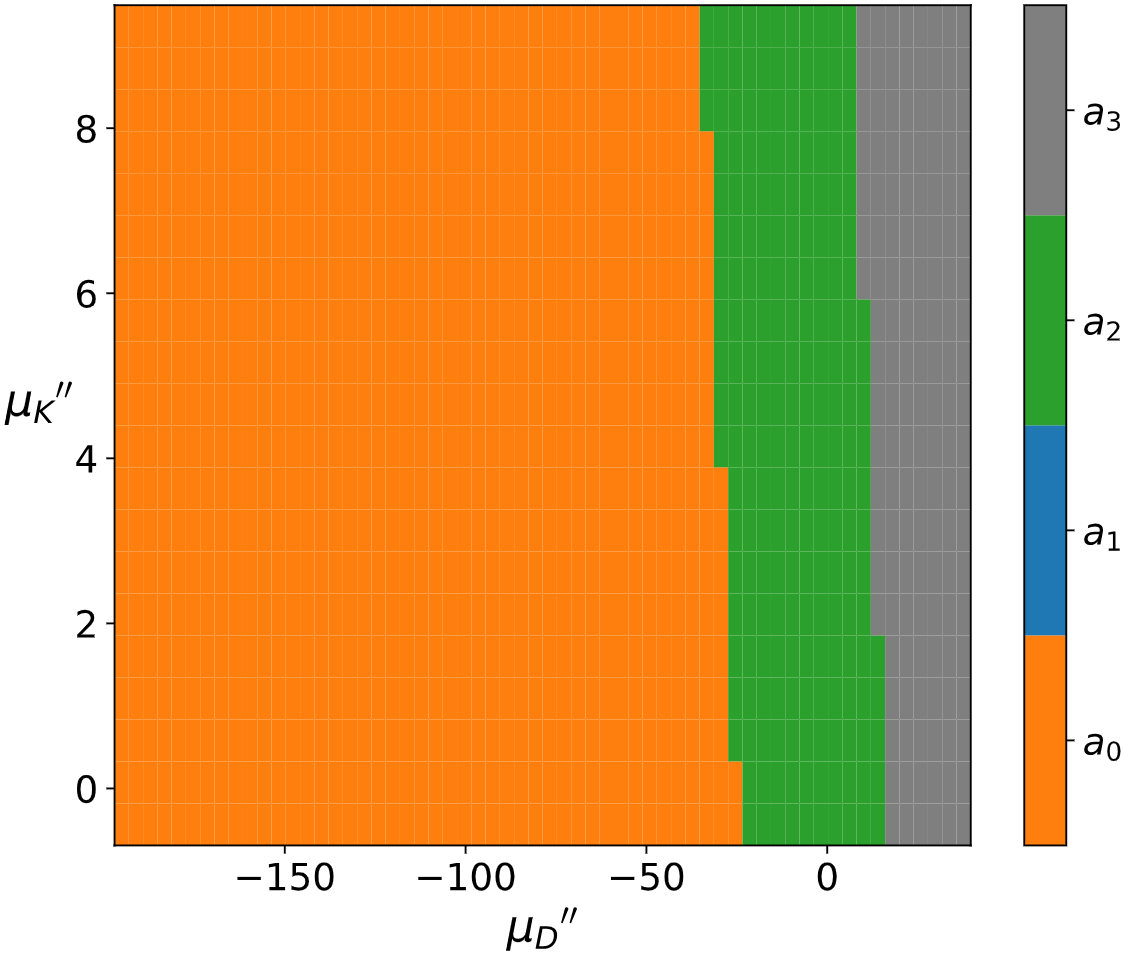}
    }
    %\captionsetup{width=0.95\textwidth}
    \caption{Evolution of the VI optimal actions represented on a $60\times20$ belief grid for $t=1$ (a), $t=10$ (b), $t=18$ (c), and $t=20$ (d) for an observation error of $\sigma_E = 50$, where the cell mid-points are chosen as representatives for each cell region, respectively.}
    \label{fig:VI_belief_grid}
\end{figure}

To investigate and compare the resulting policies, we illustrate how the strategies manifest in the belief space. \Cref{fig:VI_belief_grid} depicts the policies resulting from VI for the reference case of $\sigma_E=50$ and $t=\{1,10,18,20\}$. The occasional islands in otherwise continuous action bands in panels (a) and (b) result from the sampling-based estimation of the belief transition probabilities outlined in \citep{straub2009stochastic}. 

\begin{figure}[H]
    \centering
    \captionsetup[subfigure]{position=t, justification=centering}
    \subfloat[t=1]{
    \includegraphics[width=0.46\textwidth]{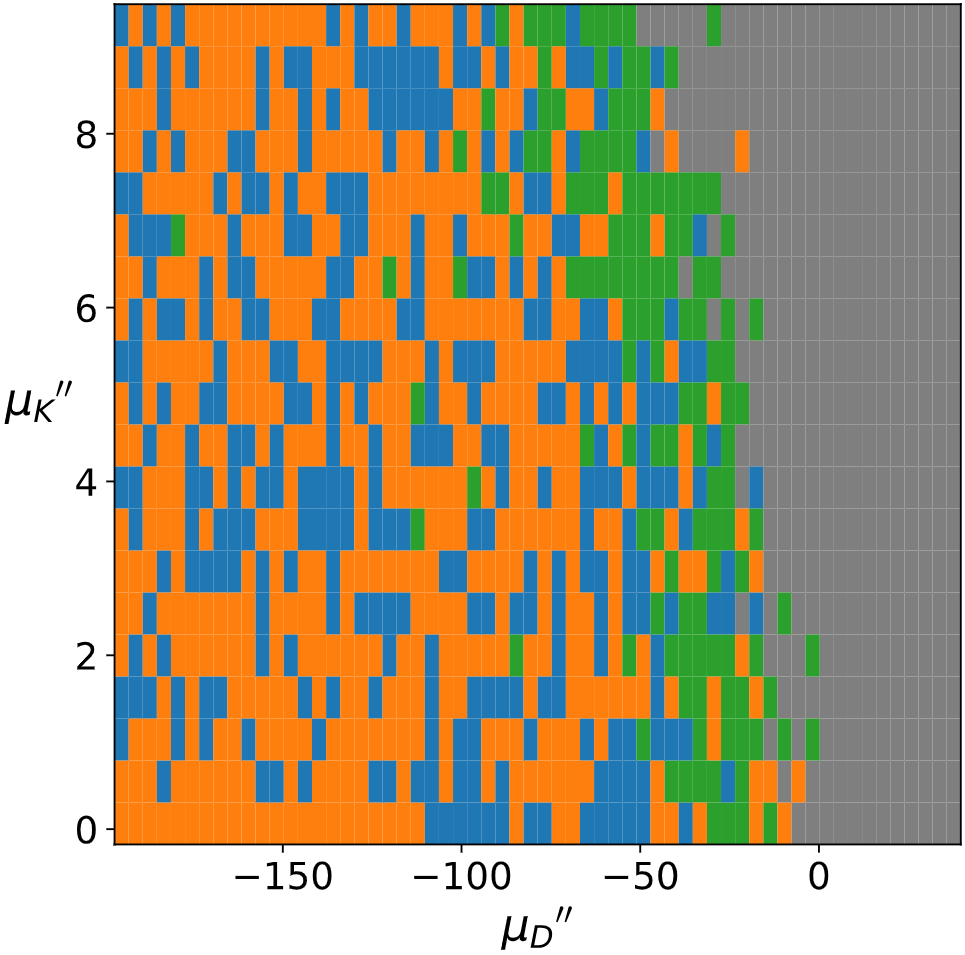}
    }
    \subfloat[t=10]{
    \includegraphics[width=0.54\textwidth]{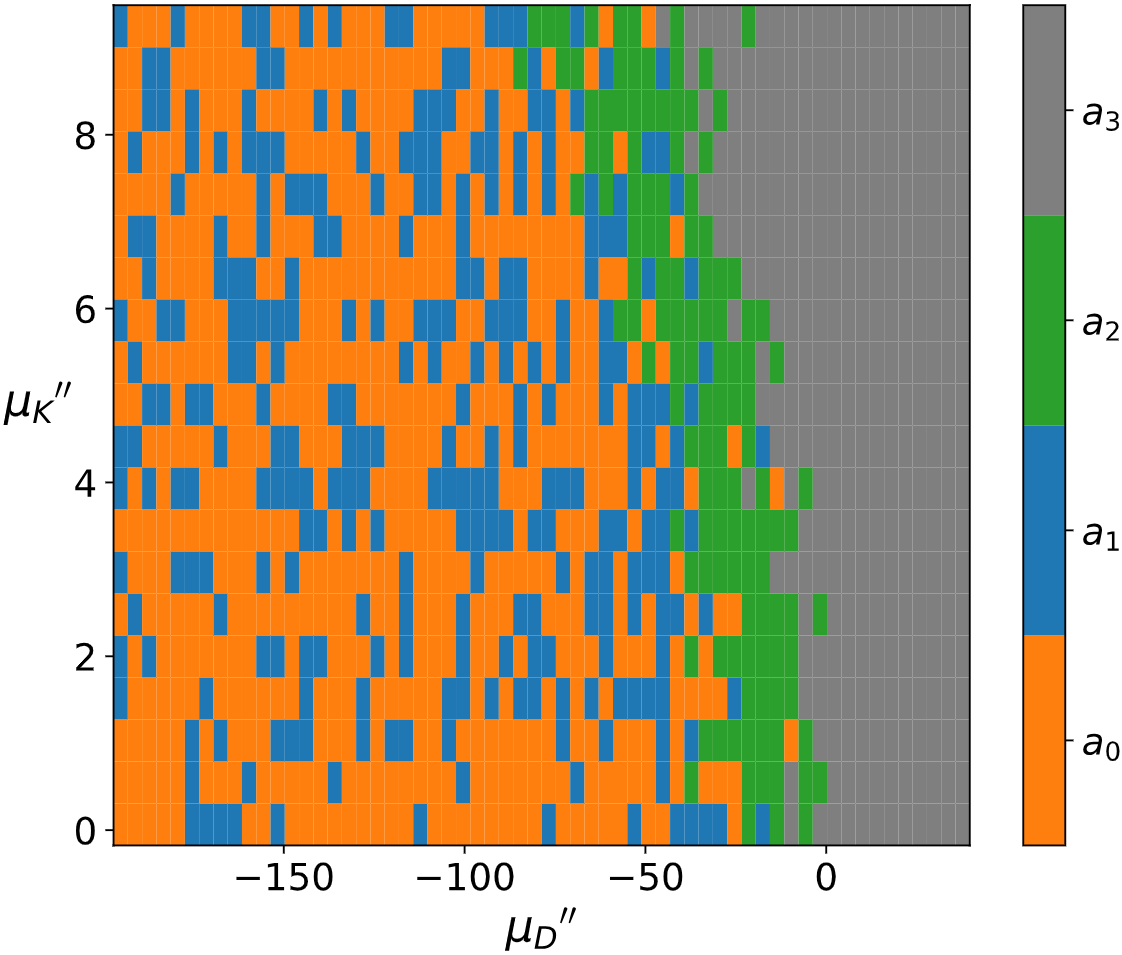}
    }
    \newline
    \subfloat[t=18]{
    \includegraphics[width=0.46\textwidth]{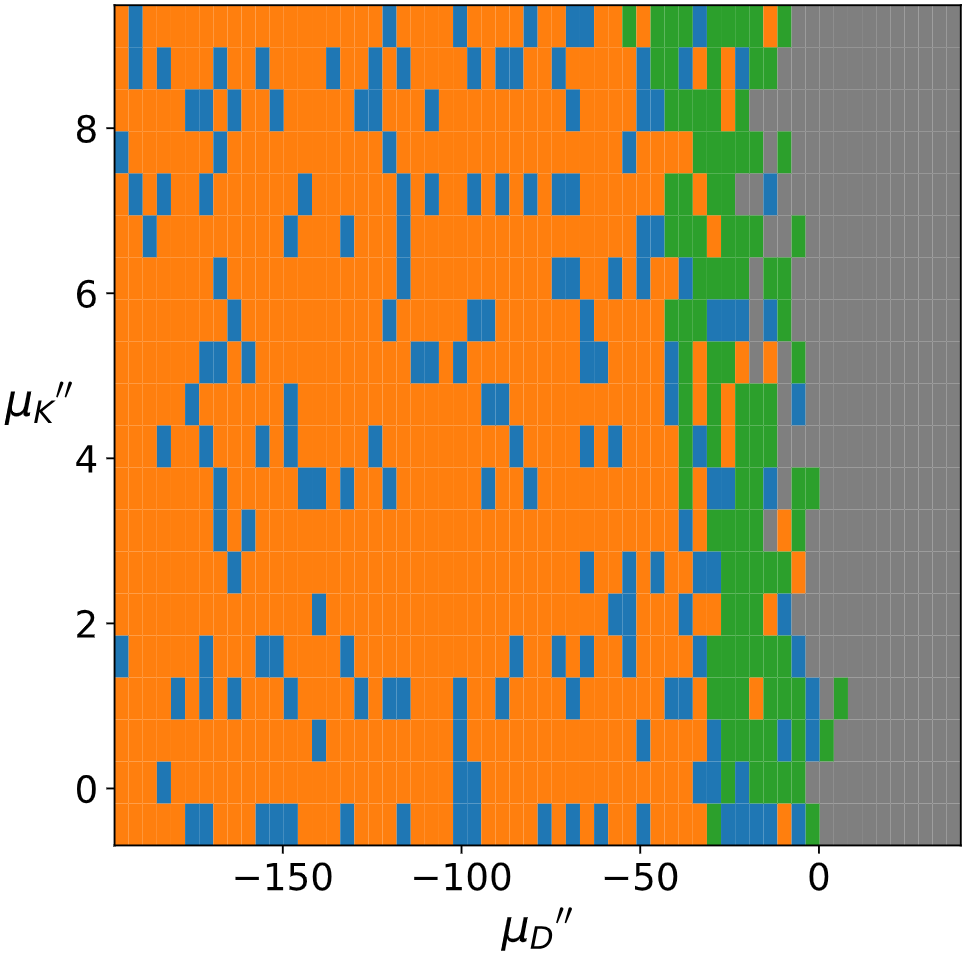}
    }
    \subfloat[t=20]{
    \includegraphics[width=0.54\textwidth]{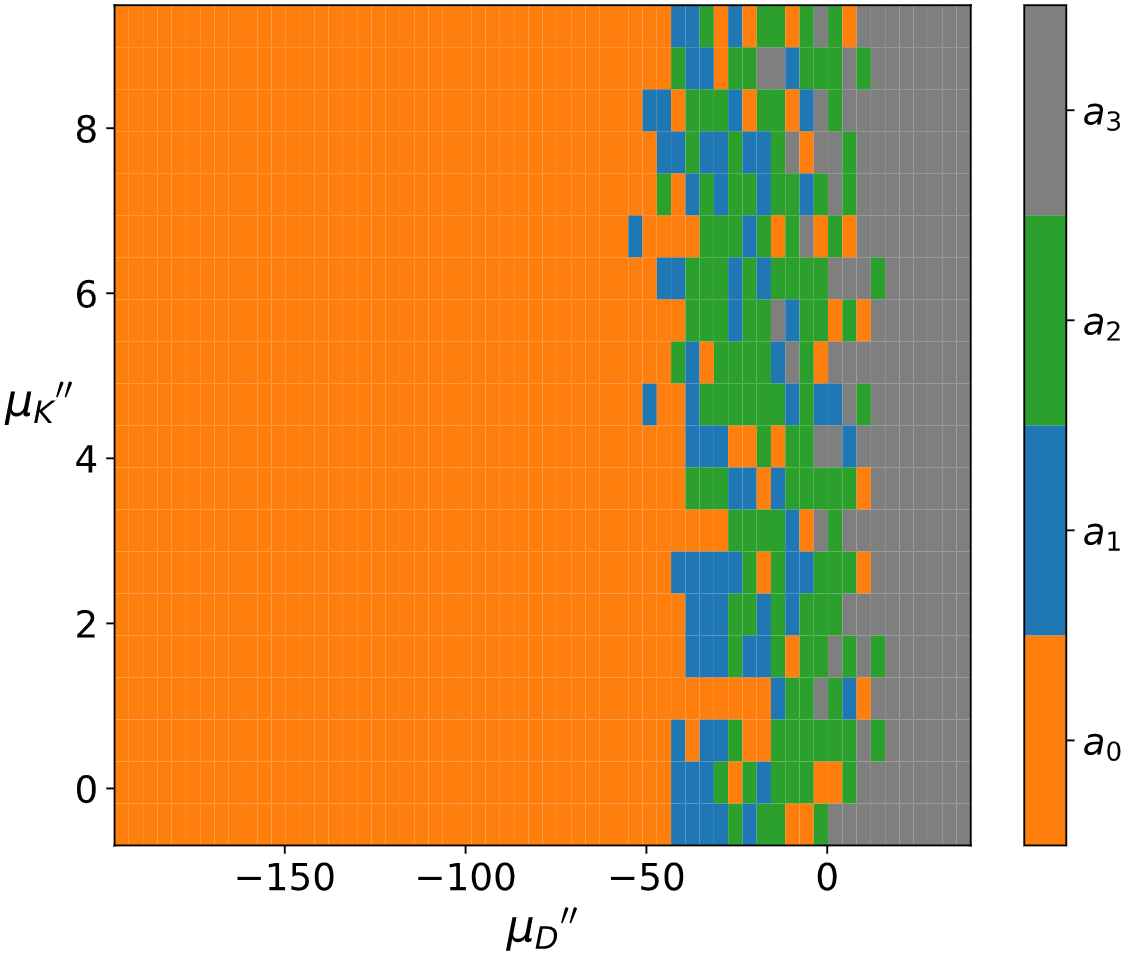}
    }
    %\captionsetup{width=0.95\textwidth}
    \caption{Evolution of the MCTS recommended actions, same set-up as in \Cref{fig:VI_belief_grid}.}
    \label{fig:MCTS_belief_grid}
\end{figure}

For comparison, we show the output of one run of the MCTS method (one for each $\vect{b}$ and $t$) in \Cref{fig:MCTS_belief_grid}. The policies are similar to the VI policies in the choice of $a_2$ and $a_3$, i.e., the regions close to or beyond failure are primarily occupied with strips of $a_2$ and $a_3$. The extent of variation is determined by the magnitude of the measurement error as well as the remaining time until the end of the life cycle, e.g., almost no variation for $\sigma_E<<1$ \& $t>10$, and high variation with no apparent structure for $\sigma_E>100~\forall t$. By contrast, the region far away from failure almost always shows high variability, and it seems that the choice between actions $a_0$ and $a_1$ is taken more or less randomly (except for very low $\sigma_E$ at $t=20$). The already mentioned variation tendencies for $a_2$ and $a_3$ also hold for $a_0$ and $a_1$. The large variance of MCTS (which could be reduced with increasing computational cost, see \Cref{subsec:mcts_optimization_technique}) leads to suboptimal policies. 

\begin{figure}[H]
    \centering
    \includegraphics[width=0.95\textwidth]{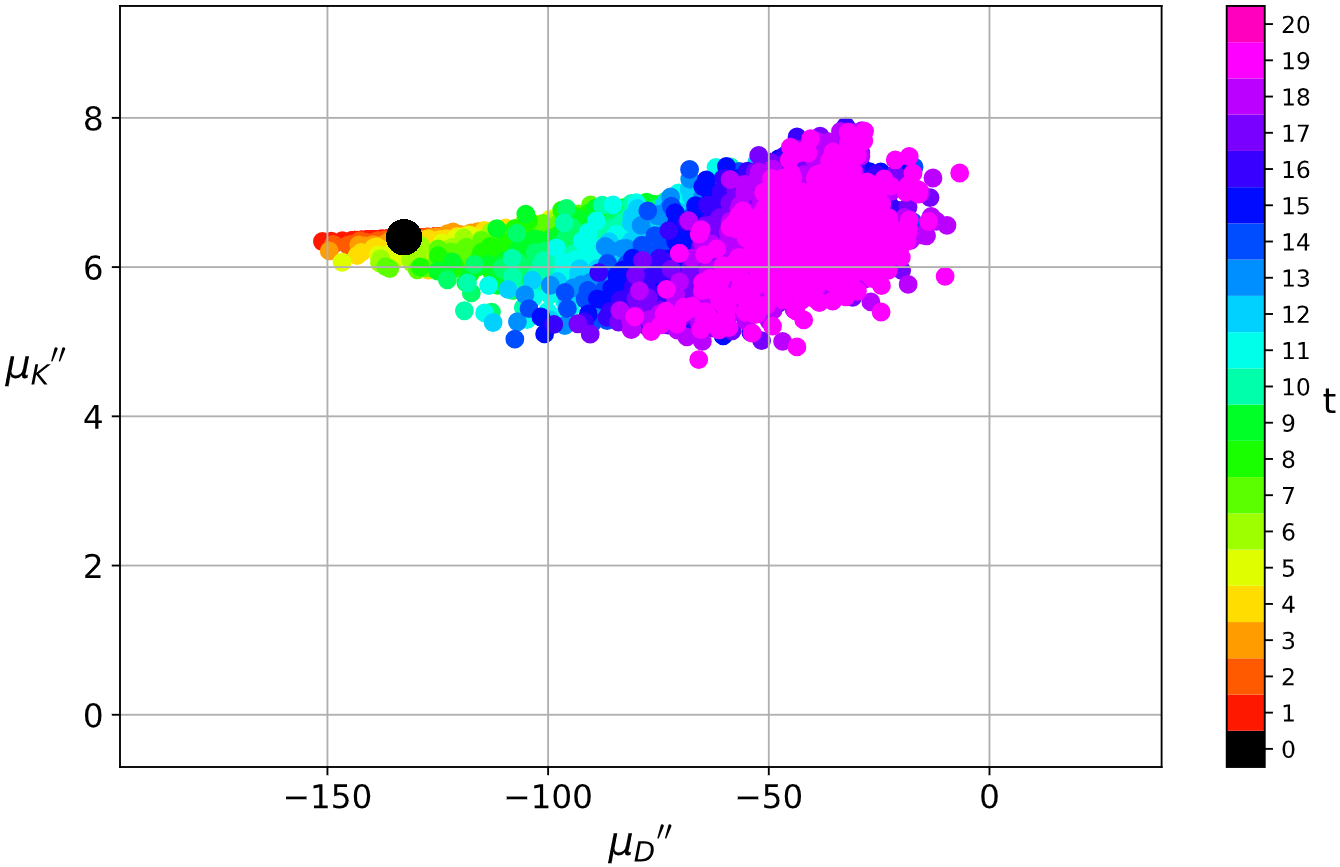}
    %\captionsetup{width=0.95\textwidth}
    \caption[Batch of belief trajectories for $\sigma_E=50$]{Evolution of a batch of 500 trajectories through time, where each color represents a specific timestep according to the colorbar; generated with the NN trained on $\sigma_E=50$.}
    \label{fig:batch_belief_trajectory_sigma_E_50}
\end{figure}

\begin{figure}[H]
    \centering
    \includegraphics[width=0.95\textwidth]{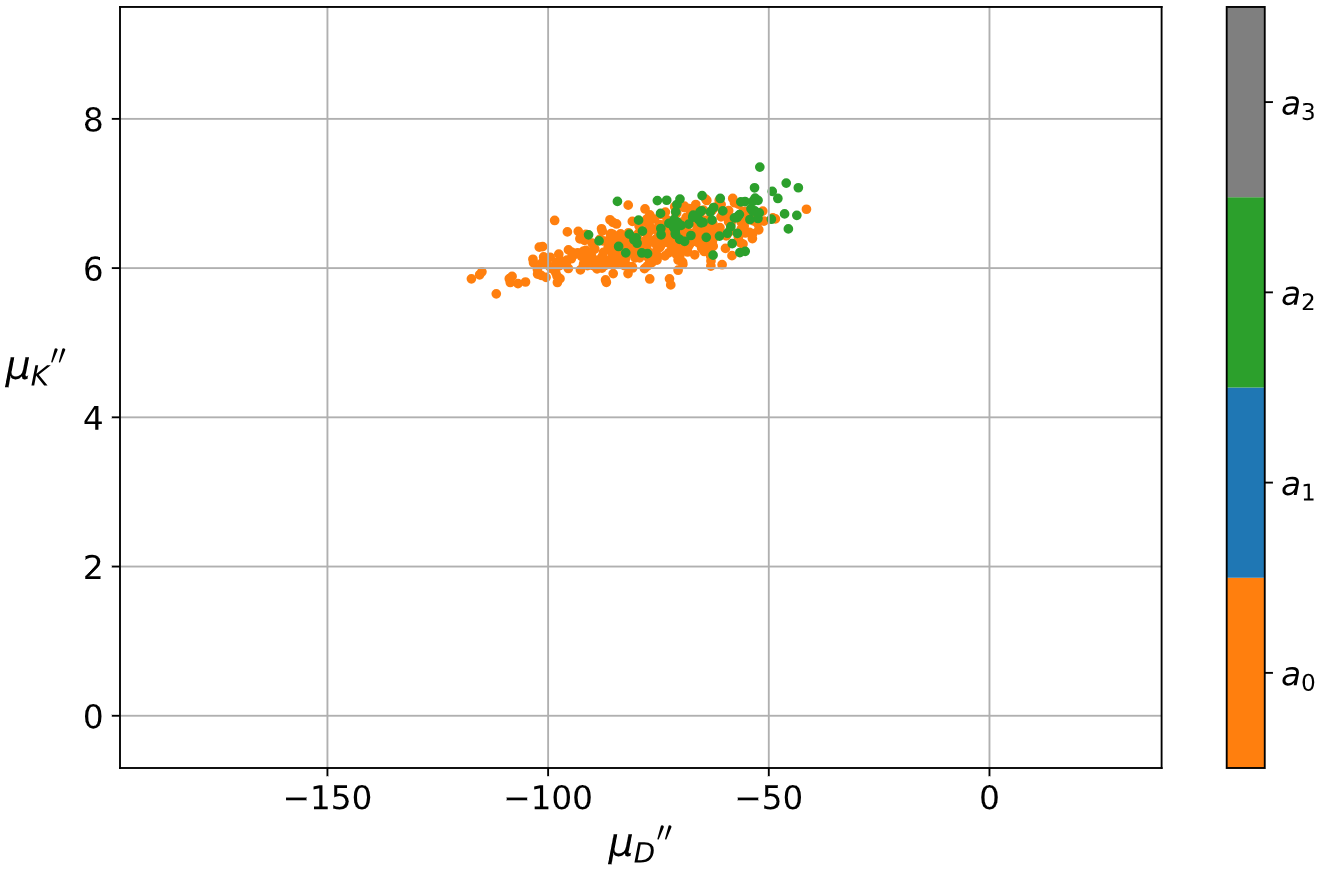}
    %\captionsetup{width=0.95\textwidth}
    \caption[NN snapshot of suggested actions in belief space for $\sigma_E=50$]{NN snapshot of the suggested actions in the belief space for $\sigma_E=50$ and 500 sample points at $t=10$.}
    \label{fig:NN_time_slice_of_action_vs_belief}
\end{figure}

For the NN, mapping all belief states to the optimal actions is not straightforward, as it takes observations and not beliefs as an input. However, the belief state can be tracked over time for sample trajectories, as shown in \Cref{fig:batch_belief_trajectory_sigma_E_50}.
Once the trajectories in the belief space are available (\Cref{fig:batch_belief_trajectory_sigma_E_50}), we can select a specific timestep and plot the actions taken by NN. This results in a point cloud in the belief space, which is shown in \Cref{fig:NN_time_slice_of_action_vs_belief}.
%
%
\begin{comment}
%
\begin{itemize}
    \item graph of achieved expected LCCs against measurement errors (fig. 26) and table with corresponding standard deviations (tab. 8)
    \item show action statistics for NN and MCTS for low, mid and high $\sigma_E$ (side by side?) (figs. 27-30)
    \item show evolution of trajectories in belief space through time (figs. 34 \& 35)
    \item show evolution of MCTS' policy in the belief space (Fig. 31) + implications (irreducible stochastic solution, discrete solution, approaching ideal solution with nearing horizon)
    \item show suggested policy of NN at a specific timestep + implications (only covers visited region, deterministic in state space but stochastic in belief space)
\end{itemize}
%
\end{comment}
%
%
%
%%%%%%%%%%%%%%%%%%%%%%%%%%%%%%%%%%%%%%%%%%%%%%%%%%%%
%%%%%%%%%%%%%%%%%%%%%%%%%%%%%%%%%%%%%%%%%%%%%%%%%%%%
%%%%%%%%%%%%%%%%%%%%%%%%%%%%%%%%%%%%%%%%%%%%%%%%%%%%
%
%
%
%
%
%
\section{Discussion}
\label{sec:Discussion}
In this work, we develop a tailored NN architecture for solving the sequential decision making problem associated with maintenance of a component subject to deterioration. We evaluate its performance on a single component maintenance problem with continuous state space. We also compare the performance of an MCTS approach on this example. 

There are several deep reinforcement learning approaches in the literature, some of which also solve the optimal inspection and maintenance problem for systems with many components \citep{andriotis2019managing,zhou2022reinforcement,andriotis2021deep,saifullah2022deep,morato2022inference}. These approaches work on discrete state spaces and compute the belief to translate the problem into a Markov decision process. The motivation for investigating the comparably simpler problem in this paper is our interest in approaches that work with continuous state spaces without a belief (even if the problem that we consider actually has an easily tractable belief, which facilitates the evaluation of the algorithms in our investigation). 

The results of our investigation show that both the NN architecture as well as the MCTS approach perform suboptimally compared to the reference solution found by value iteration. It is possible to improve the performance of both approaches, in the case of NN, by additional training and hyperparameter tuning, and in the case of MCTS, by employing a larger number of samples. However, our results reflect an honest assessment of the capabilities of these methods.

The NN's solution highly depends on the local minimum found during training. This explains the non-smooth standard deviation curve and the large differences of resulting statistical strategy profiles in different training runs, as reflected in \Cref{fig:measurement_error_evolution_std}. Generally, the NN's strategy profile changes considerably for $\sigma_E>100$ as the NN approaches the solution for the case of uninformative observations. 

As evidenced by \Cref{fig:NN_time_slice_of_action_vs_belief}, the NN's policy is stochastic in the belief space, although it is deterministic in the observation space. As the optimal policy is deterministic in the belief space (as given by the VI solution shown in \Cref{fig:VI_belief_grid}), one can observe that the trained NN is not yet able to capture implicitly the underlying belief space, which is one reason for its suboptimality.  
%Hence, there is no clear deterministic mapping from the belief space to the action space, which renders policy extraction as nontrivial. 
Thus, if the belief can be computed, it should be used as an input to the NN, as this will strongly enhance its performance (see, e.g., \citep{giacomo}) and facilitate interpretability.

The MCTS provides suboptimal but still decent results for $\sigma_E\leq50$, where it trades $\mean{\mathrm{LCC}}$ for lower variance. This can also be seen in \Cref{fig:action_statistics_comparison}, where the NN employs only $a_2$ leading to an overall lower mean cost but higher variance due to the acceptance of occasional failures. For higher observation errors, the MCTS performance decreases significantly, showing a limited ability to handle uninformative observations. This is exemplified by slightly changing strategy profiles with increasing $\sigma_E$ in \Cref{fig:action_statistics_comparison}.
Interestingly, \Cref{fig:MCTS_belief_grid,fig:VI_belief_grid} show that the MCTS' general solution is similar to the VI optimal solution provided. However, the inherent stochasticity of the method results in a stochastic policy in the belief space. This property is most apparent at the beginning of the life cycle, where the long-term effects of some actions are difficult to estimate.

A disadvantage of the MCTS approach is that it has no memory; thus, each sample trajectory has to be computed independently and expensively.  
By contrast, NNs, once trained, contain all the information in the weights, and the evaluation can be performed swiftly. 

Overall, and possibly expected, the neural networks are the preferred choice. However, there are
numerous opportunities for further enhancements of both solution approaches. 

The performance investigation of the NN could be extended, for example, by studying its dependence on the network size, its generalization capabilities (e.g., increased lifetime, different distributions), or by using the belief as an input instead of the observations for comparison. Moreover, the NN architecture can be extended by incorporating a double deep Q-network (DDQN) or by replacing the LSTM architecture with transformers (see, e.g., \citep{giacomo,hettegger}).

The MCTS method could be extended by, e.g., using erroneous observations instead of exact beliefs \citep{silver2010pomcp} for performance comparison or by switching to continuous state MCTS to dispense with discretization.
NN and MCTS can also be combined by adding a planning step to the NN-based solution.

%The degradation model could be extended by, e.g., adding a random process to the transitions of $D$ and $K$ (i.e., model $\Delta d$ and $\Delta k$ as RVs), or by incorporating multiple components. 

\begin{comment}
%
\vspace{1em}
\noindent
Neural network variations:
%
\begin{itemize}
    \item Investigation of dependence on network size, e.g., find minimum number of parameters needed to achieve minimal LCC
    \item Belief as inputs instead of observations for performance comparison
    \item Incorporation of double deep Q-network (DDQN)
    \item Replace LSTM with transformers
    \item Test generalization capabilities, e.g., increase the lifetime, different distributions, etc.
\end{itemize}

\vspace{1em}
\noindent
MCTS variations:
\begin{itemize}
    %\item [\textbf{MCTS variations:}]
    \item Use erroneous observations instead of exact beliefs (as in \citep{silver2010pomcp})
    \item Switch to continuous state MCTS to dispense with discretization 
\end{itemize}

\vspace{1em}
\noindent
Model variations \& extensions:
\begin{itemize}
    %\item[\textbf{Model variations \& extensions}]
    \item Add a random process to the transitions of $D$ and $K$, e.g., by modeling $\Delta d$ and $\Delta k$ as random variables 
    \item Extend model to a multi-component system 
\end{itemize}
\end{comment}

%
%
%
%%%%%%%%%%%%%%%%%%%%%%%%%%%%%%%%%%%%%%%%%%%%%%%%%%%%
%%%%%%%%%%%%%%%%%%%%%%%%%%%%%%%%%%%%%%%%%%%%%%%%%%%%
%%%%%%%%%%%%%%%%%%%%%%%%%%%%%%%%%%%%%%%%%%%%%%%%%%%%
%
%
%
%
%
%
\section{Conclusion}
\label{sec:Conclusion}
In this work, we propose the +RQN architecture for POMDP and I\&M planning, which requires merely the erroneous observations and the previous action taken as an input. The resulting neural networks are computationally fast and achieve good performance for measurement errors over several magnitudes through policy adaption. However, NNs, in general, inherently suffer from interpretation difficulties. The trained model consists already for small problems of thousands of weights. Interpreting the results or gaining underlying physical insights and properties of the system is non-trivial. This characteristic is evident in policy extraction, which is challenging to conduct in the belief space, as beliefs cannot be imposed but only tracked along the NN's trajectories. 

By contrast, computing many histories with the MCTS method is computationally much slower. In addition, it is inherently based on constructing a tree that exponentially grows with increasing depth, which needs large amounts of memory. The results of the MCTS are comparable to the NNs for small to medium observation errors. However, for high observation errors, the MCTS method fails to adapt its policy and achieves significantly worse results compared to the NNs and VI. The key advantage of the MCTS method lies in the evaluation of their policies. Any belief combination can be specified as a starting point which greatly facilitates the interpretation of the results. 
%

%
%
%
%
%
%%%%%%%%%%%%%%%%%%%%%%%%%%%%%%%%%%%%%%%%%%%%%%%%%%%%
%%%%%%%%%%%%%%%%%%%%%%%%%%%%%%%%%%%%%%%%%%%%%%%%%%%%
%%%%%%%%%%%%%%%%%%%%%%%%%%%%%%%%%%%%%%%%%%%%%%%%%%%%
%
%
%
%
%
%
\section*{Declarations}

\subsection*{Availability of data and materials}
The environment used and/or analysed during the current study are available from the corresponding author on reasonable request.

\subsection*{Competing interests}
The authors declare that they have no competing interests

\subsection*{Funding}
This research did not receive any specific grant from funding agencies in the public, commercial, or not-for-profit sectors.

\subsection*{Authors' contributions}
\begin{itemize}
    \item Daniel Koutas: Conceptualization;  Investigation;
          Methodology; Software; Visualization; Writing - original draft, review \& editing.
    \item Elizabeth Bismut: Conceptualization; Methodology; Supervision;            Validation; Software; Writing - original draft, review \& editing.
    \item Daniel Straub: Conceptualization; Supervision; Validation; Writing - review \& editing.
\end{itemize}

%
%
%
%
%
%%%%%%%%%%%%%%%%%%%%%%%%%%%%%%%%%%%%%%%%%%%%%%%%%%%%
%%%%%%%%%%%%%%%%%%%%%%%%%%%%%%%%%%%%%%%%%%%%%%%%%%%%
%%%%%%%%%%%%%%%%%%%%%%%%%%%%%%%%%%%%%%%%%%%%%%%%%%%%
%
%
%
%
%
%
%
%\begin{appendices}
%\newpage
\appendix
\counterwithin{figure}{section}
\counterwithin{table}{section}

\label{appendix}
%
%
%
%%%%%%%%%%%%%%%%%%%%%%%%%%%%%%%%%%%%%%%%%%%%%%%
\section{Model Info}
\label{sec:app_model_info}
\subsection{Model data}
\label{sec:app_model_data}
The specific parameters for the model used in this work are outlined in \Cref{tab_summary_of_new_params}.
\begin{table}[H]
\renewcommand{\arraystretch}{1.5}
    \centering
    \begin{tabular}{|c c | c c|}
    \hline
    \multicolumn{2}{|c}{Model} & \multicolumn{2}{c|}{Costs} \\ 
    \hline
    Parameter & Value & Parameter & Value\\
    \hline
    \hline
    $\mu_{D_0}$ & $-132.64$  & $c_{a_0}$ & $0$ \\
    $\mu_{K_0}$ & $6.4$  & $c_{a_1}$ & $1$ \\
    $\sigma_{D_0}$ & $20.85$ & $c_{a_2}$ & $5$ \\
    $\sigma_{K_0}$ & $1$ & $c_{a_3}$ & $100$ \\
    $\Delta d$ & $10.5$  & $c_F$ & $150$\\
$\Delta k$ & $0.2$ & $\gamma$ & $\frac{1}{1.02}$ \\
    \hline
    \end{tabular}
    \captionsetup{justification=centering}
    \caption{Summary of model and cost parameters.}
    \label{tab_summary_of_new_params}
\end{table}
%
%
%
%%%%%%%%%%%%%%%%%%%%%%%%%%%%%%%%%%%%%%%%%%
%
\subsection{Effect of actions}
\label{subsec:app_effect_of_actions}
The (belief) state of the system is influenced by the four available actions, whose effects are detailed in \Cref{tab:action_effect_on_samples_and_belief}.
\begin{table}[t]
\renewcommand{\arraystretch}{1.5}
    \centering
    \begin{tabular}{|c|c|c|c|}
    \hline
        action & effect & state level & belief level \\
        \hline
        \multirow{2}{*}{$a_0$} & do & $D_t = D_{t-1} + K_{t-1}$ & $\mu_{D_t}^{\prime } = \mu_{D_{t-1}}'' + \mu_{K_{t-1}}''$ \\[-0.7em]
        & nothing & $K_t = K_{t-1}$ & $\mu_{K_t}^{\prime } = \mu_{K_{t-1}}''$\\
        \multirow{2}{*}{$a_1$} & reduce & $D_t = D_{t-1} + K_{t-1} - \Delta k$ & $\mu_{D_t}^{\prime } = \mu_{D_{t-1}}'' + \mu_{K_{t-1}}'' - \Delta k$\\[-0.7em]
        & det. rate & $K_t = K_{t-1} - \Delta k$ & $\mu_{K_t}^{\prime } = \mu_{K_{t-1}}'' - \Delta k$\\
        \multirow{2}{*}{$a_2$} & improve & $D_t = D_{t-1} + K_{t-1} - \Delta d$ & $\mu_{D_t}^{\prime } = \mu_{D_{t-1}}'' + \mu_{K_{t-1}}'' - \Delta d$\\[-0.7em]
        & det. state & $K_t = K_{t-1}$ & $\mu_{K_t}^{\prime } = \mu_{K_{t-1}}''$\\
        \multirow{2}{*}{$a_3$} & replace & $D_t = \tilde{D}_{0} + \tilde{K}_{0}$ & $\mu_{D_t}^{\prime } = \mu_{D_{0}}'' + \mu_{K_{0}}''$ \\[-0.7em]
        & system & $K_t = \tilde{K}_0$ & $\mu_{K_t}^{\prime } = \mu_{K_{0}}''$\\
        \hline
    \end{tabular}
    %\captionsetup{width=\textwidth}
    \caption[Effect of actions on states and beliefs]{Mathematical description of the action $a_i$ effects on individual $D$ and $K$ states as well as their corresponding beliefs $\mu_{D}$ and $\mu_K$. At $a_3$, the replacement is conducted by drawing new samples $\tilde{D}_0$ and $\tilde{K}_0$ (from the distribution in \Cref{eq_a3_sample_distribution}) and not by reusing the samples from the current simulation.}
    \label{tab:action_effect_on_samples_and_belief}
\end{table}

Action $a_3$ consists of sampling new values for the deterioration state and deterioration rate from the following multivariate normal distribution:
\begin{linenomath}
\begin{equation}
\label{eq_a3_sample_distribution}
\left.
    \begin{bmatrix}
    D_t \\ K_t
    \end{bmatrix}
\right|_{a_{3,t-1}}
    \sim \mathcal{N}_2 \left( 
    \begin{bmatrix}
    \mu_{D_{0}} + \mu_{K_{0}} \\ 
    \mu_{K_{0}}
    \end{bmatrix}, ~
     \begin{bmatrix}
    \sigma_{D_{t}}^{\prime  2} & 
    \rho_t^{\prime } \sigma_{D_{t}}^{\prime } \sigma_{K_{t}}^{\prime } \\
    \rho_t^{\prime } \sigma_{D_{t}}^{\prime } \sigma_{K_{t}}^{\prime } &
    \sigma_{K_{t}}^{\prime  2}
    \end{bmatrix}
    \right),
\end{equation}
\end{linenomath}
where we denote with "$'$" and "$''$" the prior and posterior distributions, respectively. The corresponding analytical terms are detailed in the following.
%
%
%%%%%%%%%%%%%%%%%%%%%%%%%%%%%%%%%%%%%%%%%%%%%
%
\subsection{Transition probabilities - state level}
\label{subsubsec:app_transition_probabilities_state_level}
At every timestep $t\geq1$, after observing $O_t$, the updated distribution of $D_t$ and $K_t$ is a binormal distribution, with mean $\mu_{D,t}''$, $\mu_{K,t}''$, standard deviations $\sigma_{D,t}''$, $\sigma_{K,t}''$ and correlation coefficient $\rho_t''$.
%
%
%
%%%%%%%%%%%%%%%%%%%%%%%%%%%%%%%%%%%%%%%%%%%%%%%%%%%%%%
%
\subsubsection{Prior and posterior covariance matrix of $D_t$ and $K_t$}
\label{subsubsec:app_posterior_covariance}
For the covariance matrix, the transition from $''_{t-1}$ to $''_{t}$ does not depend on $O_t$ or $A_t$, hence is deterministic:
\begin{linenomath}
\begingroup
\allowdisplaybreaks
\begin{align}
\label{eq_sigma_D_t_prior}
\sigma_{D, t}' &=\sqrt{\sigma_{K, t-1}^{\prime \prime 2}+\sigma_{D, t-1}^{\prime \prime 2}+2 \rho_{t-1}'' \sigma_{K, t-1}'' \sigma_{D, t-1}''} \\[0.5em]
\label{eq_sigma_D_t_posterior}
\sigma_{D, t}'' &=\frac{\sigma_{E} \sigma_{D, t}'}{\sqrt{\sigma_{E}^{2}+\sigma_{D, t}^{\prime 2}}} \\[0.5em]
\label{eq_sigma_K_t_prior}
\sigma_{K, t}' &=\sigma_{K, t-1}^{\prime\prime} 
\\[0.5em]
\label{eq_sigma_K_t_posterior}
\sigma_{K, t}'' &=\frac{\sigma_{K, t}' \sqrt{\sigma_{E}^{2}+\sigma_{D, t}^{\prime 2}\left(1-\rho_{t}^{\prime 2}\right)}}{\sqrt{\sigma_{E}^{2}+\sigma_{D, t}^{\prime 2}}}
\\[0.5em]
\label{eq_rho_t_prior}
\rho_{t}'&= \frac{\rho_{t-1}'' \sigma_{D, t-1}''+\sigma_{K, t-1}''}{\sqrt{\sigma_{K, t-1}^{\prime \prime 2}+\sigma_{D, t-1}^{\prime \prime 2}+2 \rho_{t-1}'' \sigma_{K, t-1}'' \sigma_{D, t-1}''}} 
\\[0.5em]
\label{eq_rho_t_posterior}
\rho_{t}''&=\frac{\rho_{t}' \sigma_{E}}{\sqrt{\sigma_{E}^{2}+\sigma_{D, t}^{\prime 2}\left(1-\rho_{t}^{\prime 2}\right)}}.
\end{align}
\endgroup
\end{linenomath}
%
%
%
%%%%%%%%%%%%%%%%%%%%%%%%%%%%%%%%%%%%%%%%%%%%%%%%%%%%%%%%%
%
\subsubsection{Posterior mean values of $D_t$ and $K_t$}
\label{subsubsec:app_posterior_mean_values}
Conversely to the covariance matrix, the posterior mean values of $D_t$ and $K_t$ depend on the value of the observation $O_t$
\begin{linenomath}
\begin{align}
	\label{Eq:mu_D_post}&\mu_{D,t}''=\frac{\sigma_{D,t}''^2}{\sigma_\epsilon^2}O_t+\frac{\sigma_{D,t}''^2}{\sigma_{D,t}'^2}\mu_{D,t}'\\
	\label{Eq:mu_K_post}&\mu_{K,t}''=\frac{\rho_t'\sigma_{D,t}'\sigma_{K,t}'}{\sigma_\epsilon^2+\sigma_{D,t}'^2}(O_t-\mu_{D,t}')+\mu_{K,t}'.
\end{align}
\end{linenomath}
%
%
%
%%%%%%%%%%%%%%%%%%%%%%%%%%%%%%%%%%%%%%%%%%%%%%%%%%%%%%%%%
\subsection{Transition probabilities - belief level}
\label{subsec:app_transition_probabilities_belief_level}
The covariance of $D_t$ and $K_t$ is fully known (does not depend on $O_t$). The means of the distributions are fully observed at each timestep (see \Cref{Eq:mu_D_post,Eq:mu_K_post}). The belief $B_t$ at time $t$ is composed of the two posterior means, $\mu_{D,t}''$ and $\mu_{K,t}''$. From \Cref{Eq:mu_D_post} and $O_t|D_t\sim\mathcal{N}(D_t,\sigma_\epsilon)$, which gives $O_t\sim\mathcal{N}(\mu_{D,t}',\sqrt{\sigma_\epsilon^2+\sigma_{D,t}'^2})$, we obtain that %
\begin{linenomath}
\begin{equation}
	\mu_{D,t}''|B_{t-1},A_t \sim \mathcal{N}\left[
	    \mu_{D,t}'(B_{t-1},A_t),\frac{\sigma_{D,t}'^2}{\sqrt{\sigma_\epsilon^2+\sigma_{D,t}'^2}}
	\right].
\end{equation}
\end{linenomath}

One can show that $\mu_{K,t}''$ is fully correlated with $\mu_{D,t}''$ conditional on the belief at $t_1$:
\begin{linenomath}
\begin{equation}
    \mu_{K,t}'' = \frac{\rho_t'\sigma_{K,t}'}{\sigma_{D,t}'}(\mu_{D,t}''-\mu_{D,t}')+\mu_{K,t}'.
\end{equation}
\end{linenomath}
%

%
%
%
%%%%%%%%%%%%%%%%%%%%%%%%%%%%%%%%%%%%%%%%%%%%%
\section{Neural network specifications}
\label{sec:app_nn_specs}

%
\begin{comment}
\subsection{Architecture}
\label{subsec:app_architecture}
%
The architecture used in this work is a fusion of the works of \citep{zhu2017action-specific} and \citep{wang2016dueling}, and is depicted in \Cref{fig:snapshot_network_architecture}
%
\nnarchitecture
%
\end{comment}
%
%%%%%%%%%%%%%%%%%%%%%%%%%%%%%%%%%%%%%%%%%%%%%%%%%
%%%%%%%%%%%%%%%%%%%%%%%%%%%%%%%%%%%%%%%%%%%%%%%%%
%
%
\subsection{Fixed NN parameters}
\label{subsec:app_fixed_NN_parameters}
The number of hidden layers loosely follows the architecture from \citep{zhu2017action-specific}. It is possible that the network achieves better performance or equal performance with shorter training time with other configurations. 

The output dimensions of $O,~a,~A,~V,~\text{and}~Q$ are fixed by our problem formulation, i.e., we have one observation variable and four available one-hot encoded actions. The output dimensions of all other layers - the three FC layers and the LSTM layer - can be freely chosen. The dimensions of all customizable layers have been selected heuristically. The fully connected layers are numbered according to the order in which they appear from left to right, i.e., there are two FC1 and FC2 layers. The sizes of FC2 (and FC1) have been chosen to have the same dimension to not impose an ad hoc ranking of importance before they enter the LSTM layer. The exact values for the number of nodes in each layer and all other parameters set heuristically are given in \Cref{tab:network_and_optimizer_params}.
\begin{table}[H]
\renewcommand{\arraystretch}{1.5}
\centering
    \begin{tabular}{|c c | c c|}
    \hline
    \multicolumn{2}{|c}{Network} & \multicolumn{2}{c|}{Optimizer} \\ 
    \hline
    Parameter & Value & Parameter & Value\\
    \hline
    \hline
    FC1 size & 20 & Optim. type & Adam \citep{kingma2014adam} \\
    FC2 size & 25 & Learning rate & $0.001$ \citep{PyTorchAdam}\\
    Hidden state size & 80 & Betas & $(0.9, 0.999)$ \citep{PyTorchAdam} \\
    FC3 size & 160  &  AMSGRAD & Included \citep{reddi2019amsgrad} \\
    FC activation funcs. & Leaky ReLU & Batch size & 500 \\
    Leaky ReLU slope & 0.3 & Epochs & 500 (at most) \\
    Target update & 3 & $\epsilon$ decrease & 0.1 \\
    \hline
    \end{tabular}
    \captionsetup{justification=centering}
    \caption[Network and optimizer parameters]{Summary of new heuristically chosen network and optimizer parameters}
    \label{tab:network_and_optimizer_params}
\end{table}
The total number of parameters of our NN architecture for the specific values given in \Cref{tab:network_and_optimizer_params} is 57,195.

We train our network for at most 500 epochs. However, early stopping is also implemented, i.e., training is interrupted if the training loss does not further decrease over an extended period \citep{I2DL}.
%
%
%%%%%%%%%%%%%%%%%%%%%%%%%%%%%%%%%%%%%%%%%%%%%%%%
%%%%%%%%%%%%%%%%%%%%%%%%%%%%%%%%%%%%%%%%%%%%%%%%
%
%
\subsection{Optimized NN parameters}
\label{subsec:app_optimized_NN_parameters}
Our chosen parameters to optimize are given in the following list.
\begin{enumerate}
    \item Weight decay parameter $\lambda$ (L2 regularization) \citep{I2DL}
    \item Maximum $\epsilon$ value (coupled with a decrease)
    \item Learning rate step size
    \item Learning rate multiplication factor
\end{enumerate}
Including weight decay, the loss function gets an additional term:
\begin{linenomath}
\begin{equation}
\label{eq_full_loss_with_regularization}
\mathcal{L} = \mathcal{L}_{\mathrm{MSE}} + \lambda \cdot R(\vect{W}),
\end{equation}
\end{linenomath}
where $\vect{W}$ is a matrix containing all network weights, $R$ denotes the regularization function, which is the squared sum of all network weights ($L_2$), and $\lambda$ is a scaling parameter determining the relative importance of the regularization term compared to the MSE loss. We search for an optimal value of $\lambda$. 

We implement our behaviour policy, i.e., the policy with which we select the next action when generating a batch of trajectories, as a decreasing $\epsilon$-greedy method which starts with the value $\epsilon$ in the beginning to fuel exploration but decreases to 0 for exploitation of the final policy. However, one can also choose a different minimum $\epsilon$-value (e.g., 0.1 in \citep{zhu2017action-specific}) to always force some exploration. Our update scheme takes the form of:
\begin{linenomath}
\begin{equation}
\label{eq_eopsilon_decrease}
    \epsilon \longleftarrow \epsilon - 0.1.
\end{equation}
\end{linenomath}
Therefore, our scheme implements a simple linear reduction. The starting value of $\epsilon$ is optimized. 
%As the minimum value and decrease value are fixed (again out of heuristics), we seek to optimize the starting and thus maximum value $\epsilon$.

We also implement a learning rate scheduler, where the learning rate starts at a high value and periodically decreases, which helps both generalization and optimization \citep{you2019lrdecay}. We implement a simple step decay schedule that reduces the learning rate by a constant factor $\eta$ every constant number of epochs $m$ \citep{ge2019step}. Hence, we search for the optimal values of $m$ and $\eta$.

There are plenty more common practices for training NNs, e.g., weight initialization, batch normalization, and dropout. For most of these, we follow the default settings of PyTorch; these will not be further explained here.  
%
%
%
%%%%%%%%%%%%%%%%%%%%%%%%%%%%%%%%%%%%%%%%%%%%%%%%%%%%%%%%%%%%%%
\subsection{NN Optimization technique}
\label{subsec:NN_optimization_technique}
Several search techniques can be employed to find good NN hyperparameters. The most common one is manual search, which is simple and effective for finding reasonable estimates (e.g., initial learning rate), but becomes unstructured and ineffective when the search space of the parameters to tune grows. Therefore, we use grid search, where we define a set of points for each of our desired hyperparameters and iterate over all possible combinations \citep{I2DL}. During the procedure, we track the performances of each network and select the best-performing one.
%
%
%
%
%
%%%%%%%%%%%%%%%%%%%%%%%%%%%%%%%%%%%%%%%%%%%%%
%%%%%%%%%%%%%%%%%%%%%%%%%%%%%%%%%%%%%%%%%%%%%
\section{MCTS tuning}
\label{sec:app_MCTS_optimization}
A number of parameters influence the performance of the tree search method and hence need to be optimized. The disadvantage of MCTS compared to NNs is that parameters cannot be passed as an input, and the method finds the optimal values by itself. In addition, it takes too much time to generate an accurate representation of the performance of the tree; therefore, we cannot use an extensive grid search as we did with the NNs (\Cref{subsec:NN_optimization_technique}). Thus, we try to minimize the number of parameters we need to optimize. The remaining parameters are then analysed sequentially with appropriate assumptions.
%
%
%
%%%%%%%%%%%%%%%%%%%%%%%%%%%%%%%%%%%%%%%%%%%%%%%%%%%%%%%%%%%%%%%%%%%%%%%%%%%%%%%%%%%%%%%%%%%%%%%%%%%%%%%%%%%
\subsection{Fixed MCTS parameters}
\label{subsec:fixed_MCTS_parameters}
%
\begin{comment}
%
\begin{table}[H]
\renewcommand{\arraystretch}{1.5}
    \centering
    \begin{tabular}{|c c}
    \hline
    c & 1 \\
    $d_{fl}$ & -159.36 \\
    $d_{ce}$ & 26.67 \\
    \hline
    \end{tabular}
    \captionsetup{justification=centering}
    \caption{Summary of fixed MCTS parameters.}
    \label{tab:summary_of_fixed_mcts_params}
\end{table}
%
\end{comment}
%
The variable $c$ in \Cref{eq_uct} is also called the \emph{exploration constant}, since it expresses the weight of exploration (second term) compared to exploitation (first term). When $c=0$, one has a purely greedy policy \citep{gelly2011monte}. On the other hand, when $c \longrightarrow \infty$, one has a purely exploratory policy. We conveniently set $c=1$, but other approaches exit (see, e.g., \citep{silver2010pomcp}.

The next parameters that we fix are the upper and lower bounds of the observation buckets. The MCTS algorithm works with discrete observations, but our case study concerns a continuous deterioration and a continuous observation space. Although there exist MCTS variations which can deal with continuous action and state spaces (see, e.g., \citep{couetoux2013mctsvariation}), we can easily transform our problem to the discrete space by bucketing our observations, i.e., a certain bucket points to a range of observations. The question that now arises is how to choose these buckets. Generally, the bucket size does not have to be constant, but for simplicity, we choose buckets of equal size (with the exception of the first and last bucket). Therefore, we only need to define the ceiling ($d_{ce}$) and floor ($d_{fl}$) bound, as well as the number of desired observation buckets to fully define our buckets. The general case of $N_{ob}$ equal-sized observation buckets is depicted in \Cref{fig_bucket_bounds_general_case}: \vspace{1em}
\generalbucketbounds
Thus, we need to find some reasonable values for the floor and ceiling values $d_{fl}$ and $d_{ce}$. We can relate $d_{fl}$ to a percentile of the initial distribution of $D$, and $d_{ce}$ to a percentile of the final distribution of $D$ when letting the system evolve without intervention (i.e., the "worst" case). This leads to:
\begin{linenomath}
\begin{align}
\label{eq_d_fl}
    d_{fl} &= -159.36 
    \\[0.5em]
    d_{ce} & = 26.67.
    \label{eq_d_ce}
\end{align}
\end{linenomath}
%
%
%
%
%%%%%%%%%%%%%%%%%%%%%%%%%%%%%%%%%%%%%%%%%%%%%%%%%%%%%%%%
%
%
\subsection{Tunable MCTS parameters}
\label{subsec:tunable_MCTS_parameters}
There are further parameters that we do not set a priori but still highly influence the MCTS’ performance. The parameters to be optimized are given in the following list.  
\begin{enumerate}
    \item Tree iterations $N_T$
    \item Rollout runs $N_R$
    \item Observation buckets $N_{ob}$
\end{enumerate}
The number of tree iterations $N_T$ dictates the depth the tree reaches, i.e., the number of timesteps it looks into the future. In addition, a higher number of tree iterations increases the accuracy of the Q-value estimate. However, the possible number of nodes in our tree grows exponentially with increasing depth, and one can also increase the accuracy with the number of rollout runs $N_R$ from a given system state. Averaging over multiple rollouts instead of relying on a single run greatly reduces the susceptibility to high variances resulting from large differences in action and failure costs.

Lastly, the number of observation buckets $N_{ob}$ is also a crucial parameter, as it influences the reachable depth of the tree given a fixed number of tree iterations. In addition, it represents the degree of precision with which the observations are discretized. Hence, it is essential to find the right balance between depth and resolution.  
%
%
%
%%%%%%%%%%%%%%%%%%%%%%%%%%%%%%%%%%%%%%%%%%%%%%%%%%%%%
\subsection{MCTS parameter optimization technique}
\label{subsec:mcts_optimization_technique}
What remains now is to outline an optimization procedure for the three parameters of \Cref{subsec:tunable_MCTS_parameters} considering the observation error.
Generally, it is assumed that the optimal number of observation buckets is dependent on the observation error with regards to minimizing the LCC.  

The first analysis is conducted on the time dependence of $N_T$ and $N_R$, where we impose some threshold of computation time needed to traverse a whole life cycle with the MCTS method to stay in a computationally feasible domain. It is assumed that the computation time is independent of the observation error and is only minimally affected by the choice of $N_{ob}$, which is why they are fixed.

The result of the analysis is a set of different possible combinations of the two parameters, which satisfy our imposed computation threshold. To settle for a single combination, the influence of $N_T$ and $N_R$ on the LCC will be taken into account. We assume that the resulting curves qualitatively hold for any $N_{ob}$ and $\sigma_E$, which is why they are fixed again.

Secondly, once $N_T$ and $N_R$ have been fixed with the time constraints and LCC maximization, we search for the optimal number of observation buckets given a set of observation errors of interest.

%\end{appendices}

\bibliographystyle{elsarticle-num-names} 
\bibliography{main}





% \end{thebibliography}
\end{document}